\documentclass[preprint,1p,times]{elsarticle}

\let\today\relax
\makeatletter
\def\ps@pprintTitle{%
    \let\@oddhead\@empty
    \let\@evenhead\@empty
    \def\@oddfoot{\footnotesize\itshape
         {\empty} \hfill\today}%
    \let\@evenfoot\@oddfoot
    }
\makeatother

\usepackage{amsmath}
\usepackage{amssymb}
\usepackage{hyperref}
\usepackage{graphicx}
\usepackage{subfigure}
\usepackage{setspace}
\usepackage{bm}
\usepackage[normalem]{ulem}
\usepackage{color}
\usepackage{stfloats}
\usepackage[justification=centering]{caption}
\usepackage{setspace} 
\usepackage{array}
\usepackage{makecell}
\usepackage{changes}

\begin{document}

\begin{frontmatter}

	\title{Remaining Useful Life Prediction Using Temporal Deep Degradation Network for Complex Machinery with Attention-based Feature Extraction}
	\author[main]{Yuwen Qin}
	\author[main]{Ningbo Cai}
	\author[second]{Chen Gao}
	\author[main]{Yadong Zhang}
	\author[main]{Yonghong Cheng}
	\author[main]{Xin Chen\corref{cor1}}%
	\ead{xin.chen.nj@xjtu.edu.cn}
	\cortext[cor1]{Corresponding Author.}
	\address[main]{Center of Nanomaterials for Renewable Energy, State Key Laboratory of Electrical Insulation and Power Equipment, School of Electrical Engineering, Xi'an Jiaotong University, Xi'an 710054, Shaanxi, China.}
	\address[second]{Xi'an Thermal Power Research Institute Co.,Ltd., Xi'an 710032,China}

	\begin{abstract}
The precise estimate of remaining useful life (RUL) is vital for the prognostic analysis and predictive maintenance that can significantly reduce failure rate and maintenance costs. The degradation-related features extracted from the sensor streaming data with neural networks can dramatically improve the accuracy of the RUL prediction. The Temporal deep degradation network (TDDN) model is proposed to make the RUL prediction with the degradation-related features given by the one-dimensional convolutional neural network (1D CNN) feature extraction and attention mechanism. 1D CNN is used to extract the temporal features from the streaming sensor data. Temporal features have monotonic degradation trends from the fluctuating raw sensor streaming data. Attention mechanism can improve the RUL prediction performance by capturing the fault characteristics and the degradation development with the attention weights. The performance of the TDDN model is evaluated on the public C-MAPSS dataset and compared with the existing methods. The results show that the TDDN model can achieve the best RUL prediction accuracy in complex conditions compared to current machine learning models. The degradation-related features extracted from the high-dimension sensor streaming data demonstrate the clear degradation trajectories and degradation stages that enable TDDN to predict the turbofan-engine RUL accurately and efficiently.
	\end{abstract}




\end{frontmatter}


\section{Introduction}
The robust and accurate prediction of complex machinery's remaining useful life (RUL) is essential for condition-based maintenance (CBM) to improve reliability and reduce maintenance costs.
Turbofan engine, a key component of aircraft, is highly complex and precise thermal machinery. 
Since the degradation of turbofan engines can lead to serious disaster, it is crucial to conduct prognostics maintenance to ensure machinery safety.
Prognostics maintenance is a challenging task due to the nonlinear complexity and uncertainty in the complex machinery.
The remaining useful life (RUL) is a technical term used to describe the progression of fault in prognostics and health management (PHM) applications\cite{si2011remaining}. As a result, the PHM system can make maintenance decisions to improve the reliability, stability, and efficiency of complex machinery\cite{zhang2021transfer,pang2021bayesian}.
In particular, accurate RUL prediction is a highly crucial technology of PHM that performs the machinery's health states to provide precise failure warnings and avoid safety accidents.
Therefore, RUL prediction is essentially valuable in engineering practice.

Generally, RUL is defined as the period from the current time to the failure within a component\cite{wang2021remaining,zhang2021remaining,ellefsen2019remaining}. The existing RUL prediction algorithms can be divided into two main categories: model-based approaches and data-driven approaches.
The physical model-based approaches\cite{jouin2016degradations} are to describe the mechanical degradation process by building a mathematical model. However, the sub-components of the complex machinery are coupled, making some model parameters hard to obtain.
Building a physical model that accurately simulates the degradation process is challenging. In contrast, the data-driven approaches\cite{zhu2018estimation,ding2021remaining} do not rely on physical models but the extraction of degradation-related features from the historical data.
They can reveal the underlying correlations and causalities between raw sensor data and RUL. Therefore, the data-driven approaches attract a lot of research interests. 
The machine learning approaches utilize expert knowledge, and signal processing algorithms to extract features from raw sensor data to predict the RUL,  $\it{e.g.}$ support vector machine (SVM)\cite{ordonez2019hybrid}, hidden Markov model (HMM)\cite{chen2019hidden}, artificial neural network (ANN)\cite{gebraeel2004residual}, extreme learning machine (ELM)\cite{liu2018method}.

The machine learning approaches can reduce machinery maintenance costs because condition monitoring data can have very high dimensions. 
The underlying relationship becomes more complex, making for traditional machine learning approaches to extract temporal features from the raw sensor data. However, given the higher dimensionality and more complex relationship of raw sensor data, deep learning neural networks are more capable of RUL prediction. 
In recent years, deep learning (DL) neural networks with the ability of automatic feature extraction and great nonlinear fitting have attracted widespread attention of engineers and researchers\cite{krizhevsky2012imagenet,lecun2015deep,dahl2011context}. It has shown effectively learning capacity and achieved excellent performance, especially in image processing, natural language processing (NLP), etc. Thus, deep learning technology provides a promising solution to improve RUL prediction accuracy.
So far, the deep learning approaches, such as recurrent neural network (RNN)\cite{graves2013speech}, long short-term memory (LSTM)\cite{hochreiter1997long}, gated recurrent unit (GRU)\cite{cho2014learning} for time series modeling, as well as convolutional neural network (CNN)\cite{lecun2015deep} have been widely used for the feature extraction from high-dimensional data. 

Although the deep learning neural network has shown excellently potential for RUL prediction, some practical and data-related issues are worthy of further consideration. Since physical machinery works in complex operating conditions, sensor streaming data must contain noisy fluctuations and measurement errors. Hence, the fluctuating changes significantly affect the feature extraction and RUL prediction. Currently, expert knowledge, signal processing approaches, and health indicators are still used in feature extraction. However, these approaches cannot sufficiently mine intrinsic features from the high-dimensional sensor streaming data. There is an urgent need for an effective method to extract features reflecting the underlying physical degradation development.
Moreover, the degradation patterns in each cycle may have different weights in the degradation development, especially the representative features that appear at specific cycles in periodic pulses or shocks. If the cycle sensor data is equally considered to the RUL prediction,  the key fault characteristics are overwhelmed by noise signals. Thus, capturing key fault characteristics makes the deep learning neural network achieve highly accurate RUL prediction.

We propose the temporal deep degradation network (TDDN) combining 1D CNN with attention mechanism to achieve higher prediction accuracy by extracting the degradation-related features. Firstly, the TDDN model utilized 1D CNN to extract temporal features from raw sensor data. Next, temporal features are fed to a fully connected layer to generate abstract features, then an attention mechanism is introduced to calculate the importance weights of abstract features.  
The TDDN model outperforms the existing deep learning models on the same dataset. The main contributions of this paper are as follows:
\begin{itemize}
	\item The end-to-end TDDN model is proposed to make effective and accurate RUL prediction by extracting degradation-related features from multivariate time series.
	\item 1D CNN can accurately extract the temporal features in the degradation development. Temporal features have monotonic-degradation trends and demonstrate strong robustness against the fluctuation in the noisy streaming data.
	\item The attention weights evolve with the degradation development. Attention mechanism effectively identifies the underlying degradation patterns and captures key fault characteristics. 
	\item The effectiveness of the TDDN model on RUL prediction is verified by achieving superior performance on the C-MAPSS dataset. Furthermore, the model remarkably outperforms other methods in complex conditions on FD002 and FD004 datasets, demonstrating its robustness in practical applications.
\end{itemize}

The remaining paper is organized as follows: Section~\ref{rw} introduces related work on the C-MAPSS dataset. The proposed TDDN model is introduced and discussed in Section~\ref{TDDNs}. Section~\ref{results} illustrates the experimental approach, results, and discussions.
Section~\ref{pa} discusses the tuning of hyperparameters and how degradation-related features make the TDDN model predict RUL effectively and accurately.
We close the paper with conclusions and future work in Section~\ref{discussion}.

\section{Related Work}\label{rw}
The degradation development and machinery RUL are embedded in the streaming data collected by industrial sensors. The deep learning architectures of RNN and CNN have been widely used in RUL prediction. For the RNNs, LSTM and GRU are two deep learning architectures that are widely applied in the time series prediction. 

RNN has been widely used to mine the time series data. Heimes  $\it{et\; al.}$\cite{heimes2008recurrent} applied RNN to predict the degradation trend. Zheng  $\it{et\; al.}$\cite{zheng2017long} proposed an approach that combined LSTM cells with standard feed-forward layers to discover hidden patterns in the sensor streaming data.
The approach can take time-series temporal information into account to achieve higher accuracy.
Wang  $\it{et\; al.}$\cite{wang2019remaining}  applied LSTM to learn the nonlinear mapping from the degradation features to the RUL. These features were selected by three degradation evaluation indicators corresponding to the representative degradation features. The prediction results were superior to several traditional machine learning algorithms, such as backpropagation neural network (BPNN) and support vector regression machine (SVRM).

GRU\cite{chen2019gated} has also been applied to the RUL prediction by pre-processing sensor data with kernel principal component analysis (KPCA) for nonlinear feature extraction.
Compared with LSTM, the proposed approach took less training time and achieved better prediction accuracy.
Ahmed Elsheikh  $\it{et\; al.}$\cite{elsheikh2019bidirectional} applied the bidirectional handshaking LSTM (BHSLSTM) to process the time series streaming data in both directions effectively. This architecture enabled the LSTM network to have more insights when identifying the degradation information in time series streaming data. Chen  $\it{et\; al.}$\cite{chen2020machine} integrated the advantages of LSTM and attention mechanism to propose a deep learning framework to improve RUL prediction accuracy further. The LSTM network and attention mechanism were employed to learn temporal features and weights, respectively. A similar approach\cite{ragab2020attention} that the LSTM network encoded features and attention mechanism decoded hidden features. The encoder and decoder hidden features were fed to a fully connected neural (FCN) network for RUL prediction.
In addition, Ellefsen  $\it{et\; al.}$ \cite{ellefsen2019remaining}  introduced unsupervised deep learning techniques to overcome the difficulty of acquiring high-quality labeled training data by automatically extracting degradation features from raw unlabeled training data.
Combining the LSTM network with the restricted Boltzmann machine (RBM) network, the semi-supervised deep architecture showed better accuracy than purely supervised training approaches.

1D CNN had excellent feature learning capability to meet the challenges of highly nonlinear and multi-dimension sensor streaming data. Babu  $\it{et\; al.}$\cite{babu2016deep} adopted 1D CNN to capture the salient patterns of the sensor signals.
1D CNN was applied to learn features along temporal dimensions in the whole multivariate time series. Yao  $\it{et\; al.}$\cite{yao2021remaining} simplified the 1D CNN structure to avoid a large number of parameters by replacing the fully connected layer with the global maximum pooling layer. Then the features extracted by 1D CNN is fed into the simple recurrent unit (SRU) network to predict RUL. Li  $\it{et\; al.}$\cite{li2018remaining} improved the depth of 1D CNN. The deep 1D CNN effectively extracted degradation-related features from raw sensor data.
In addition, some researchers combined 1D CNN with other methods to make the RUL prediction. For example, the 1D CNN was combined with stacked bi-directional and uni-directional LSTM (SBULSTM) network\cite{an2020data} to learn more abstract features and encode the temporal information. Yan  $\it{et\; al.}$\cite{song2020distributed} developed an integrated model based on Temporal Convolutional Networks (TCN) with an attention mechanism to calculate the contribution of data from different sensors and degradation stages. TCN was used for feature extraction of time series, and an attention mechanism was applied to calculate temporal weights.

\section{Proposed Model: Temporal Deep Degradation Network}\label{TDDNs}
\begin{figure}[!h]
	\centering
	\includegraphics[width=1.0\textwidth]{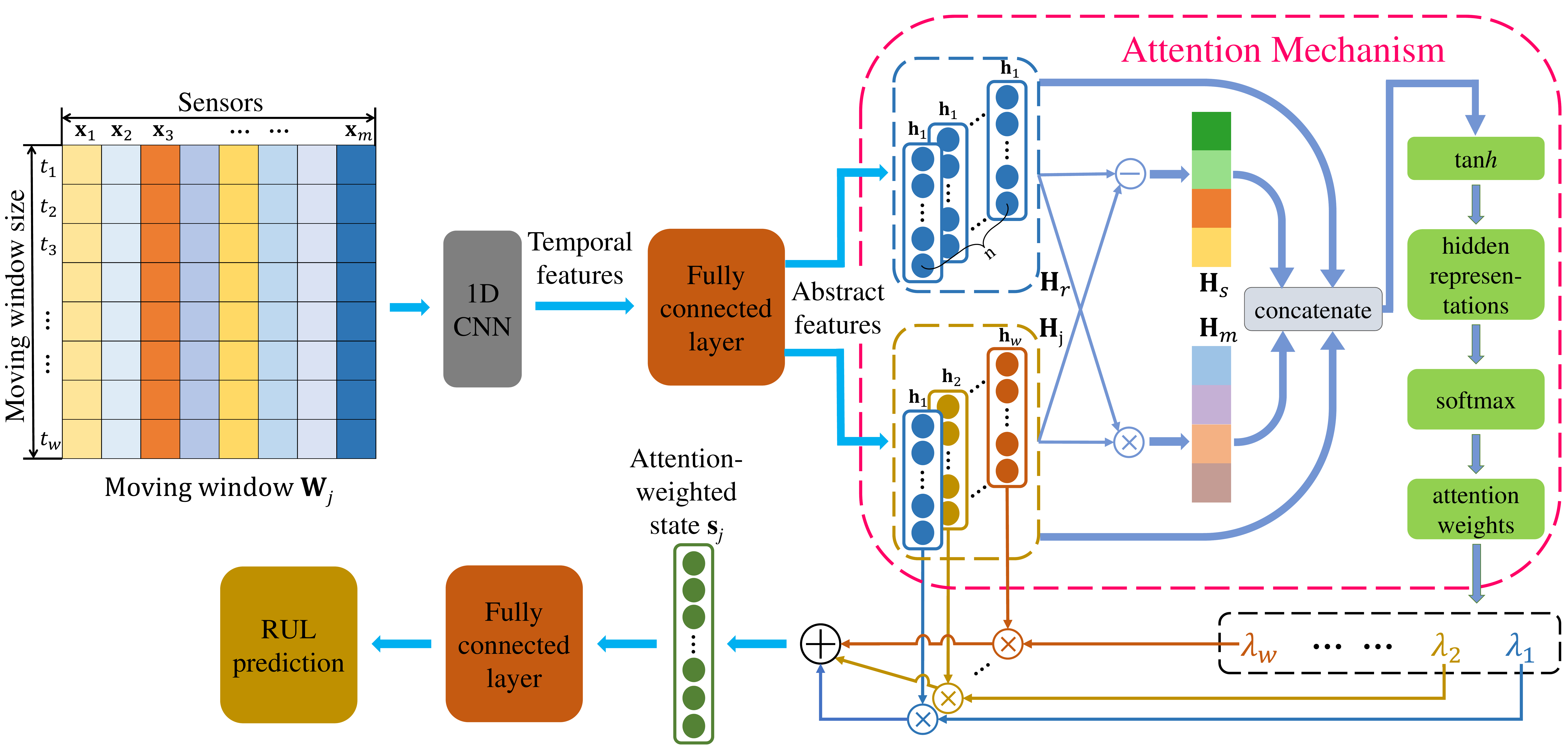}
	\caption{Proposed TDDN model with 1D CNN and attention mechanism}
	\label{TDDN}
\end{figure}

In the proposed temporal deep degradation network, 1D CNN and attention mechanism are two key components to be introduced, then the structure and parameters of the TDDN model are explained. The structure of TDDN is presented in Fig.~\ref{TDDN}, which has four parts: 1. 1D CNN to extract the temporal features, 2. fully connected layer to generate abstract features, 3. attention mechanism to generate attention-weighted state, 4. fully connected layer to predict RUL. As shown in Fig.~\ref{TDDN}, 1D CNN is firstly applied to extract temporal features from multivariate time series. Temporal features are fed to fully connected layer to generate abstract features. To capture key fault characteristics at different cycles, attention mechanism is employed to calculate a weighted sum over all abstract features. At last, the output of attention mechanism is fed into fully connected layers to predict RUL. The layer details of the proposed network are shown in Table~\ref{layerdetails}.
The 1D CNN feature extraction and attention mechanism in TDDN are critically essential for the RUL prediction. We explain them in details below.

\subsection{Sensor Data Representation and Moving Windows}
The sensor streaming data (multivariate time series) is  $\mathbf{X}=\left[\mathbf{x}_{1} ; \mathbf{x}_{2} ;  \cdots ;  \mathbf{x}_{i} ;  \cdots ; \mathbf{x}_{n}\right] \in \mathbb{R}^{n \times m}$, where $\mathbf{x}_i =\left[x_{i,1}\; {x_{i,2} \; \cdots \;  x_{i,m}}\right]$ is sensor data at time $i$,  $n$ is the length of sensor streaming data, and $m$ is the number of sensors. The sensor streaming data selection is discussed in Section~\ref{data}.
Sensor streaming data are segmented into the sequence of moving windows and used as inputs to TDDN model. With the sequence of moving windows $\mathbf{W}_M=[\mathbf{W}_1 \; \cdots \mathbf{W}_j \; \cdots \mathbf{W}_{n-w+1}]$, the $j$-th window is defined as $\mathbf{W}_j=\left[\mathbf{x}_{j} ; \mathbf{x}_{j+1} ; \cdots  ; \mathbf{x}_{j+w-1}\right]$, where $w$ is the window size.

\subsection{1D CNN and Temporal Features}
\begin{figure}[!h]
	\centering
	\includegraphics[width=\textwidth]{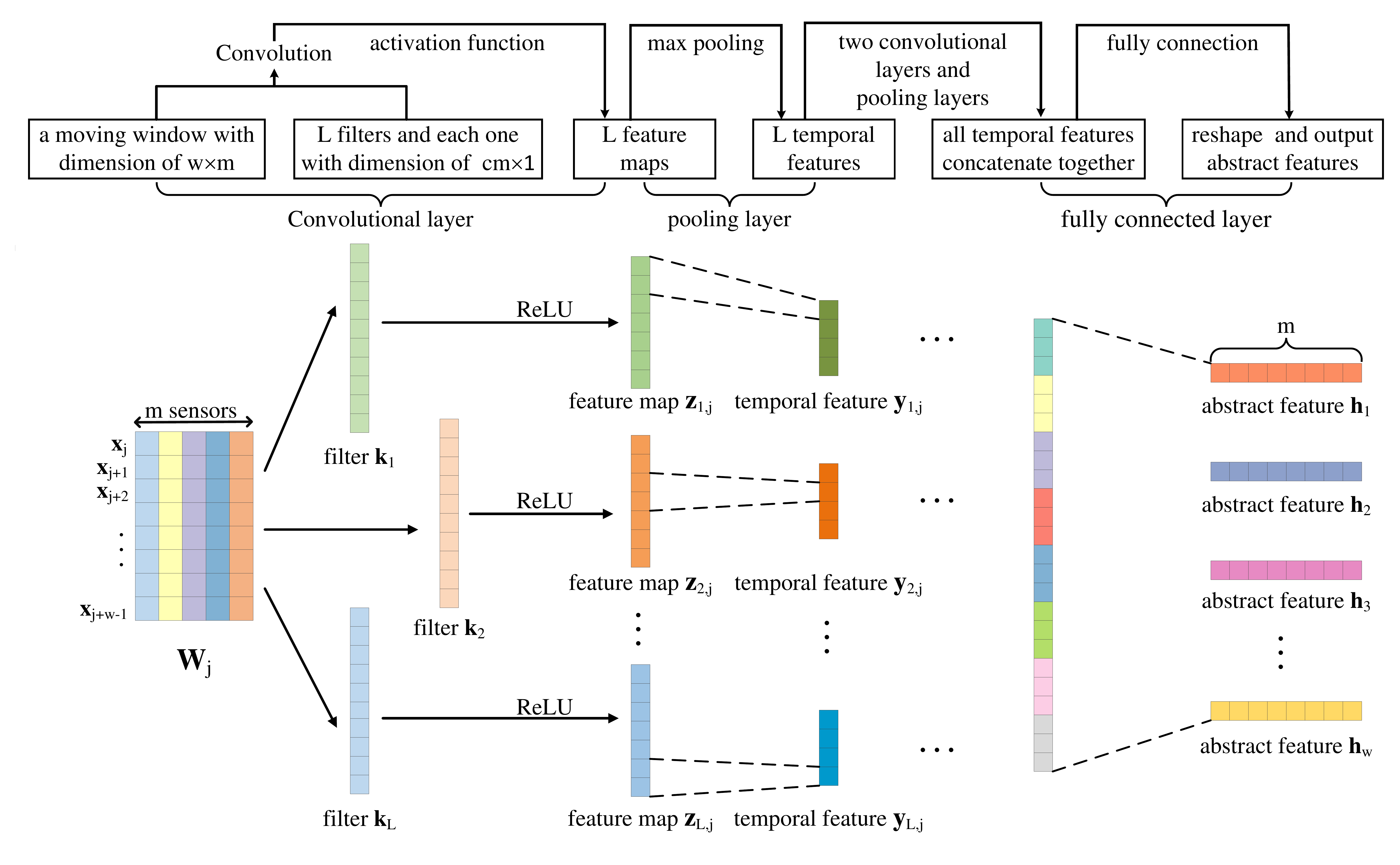}
	\caption{\rmfamily 1D  convolutional Neural Network and Temporal Feature Extraction}
	\label{1DCNN}
\end{figure}

The 1D CNN is generally composed of convolutional layers and max-pooling layers. Through the convolutional layers and max-pooling layers, the temporal features are automatically extracted from a moving window $\mathbf{W}_j$ by the filters. The 1D CNN can extract the temporal features associated with degradation patterns. The structure of 1D CNN is shown in Fig.~\ref{1DCNN}.
The convolutional filters in the convolutional layer are used to extract feature maps from the input moving window of the time series. Feature maps are fed into the pooling layer to obtain temporal features. In Fig.~\ref{1DCNN}, it is shown that $L$ filters $\mathbf{k}_l$ are used to generate $L$ feature maps. All the convolutional filters have the same dimension of $cm\times1$. Besides, the zero-padding operation is implemented in 1D CNN to ensure boundary information extraction. The rectified linear unit (ReLU) is used as the activation function for all the convolutional layers. A fully connected (FC) layer is connected 1D CNN to convert temporal features into abstract features with the same dimension $w$ as the length of the moving window.

The concatenated vector $\mathbf{x}_{k: k+c-1}^j$ of $\mathbf{x}_{j+k}$ to $\mathbf{x}_{j+k+c-1}$ in the moving window $\mathbf{W}_j$ for the 1D CNN is expressed as,
\begin{equation}
	\mathbf{x}_{k: k+c-1}^j=\mathbf{x}_{j+k} \oplus \cdots \oplus \mathbf{x}_{j+k+c-1},
\end{equation}
where $k\in [1,j+w-c]$ and symbol $\oplus$ is the direct sum of $c$ vectors. 
Therefore, the filter dimension $cm$ is equal to $c*m$.  
The $l$-th extracted feature $z_{l,j}^k$ from the concatenated vector $\mathbf{x}_{j+k: j+k+c-1}$ in the $j$-th moving window is defined as,
\begin{equation}
	z_{l,j}^k=f(\mathbf{k}_l\cdot \mathbf{x}_{k:k+c-1}^j + b_l)
\end{equation}
where
$\mathbf{k}_l$ is the $l$-th filter and
$b_l \in \mathbb{R}$ is a bias term, and $f$ and symbol $\cdot$ represent nonlinear activation function and dot product, respectively. The feature map $\mathbf{z}_{l,j}$ of the $j$-th moving window $\mathbf{W}_j$ is defined as,
\begin{equation}
	\mathbf{z}_{l,j}=\left[z_{l,j}^{1} \;\cdots  z_{l,j}^{k} \; \cdots \;  z_{l,j}^{j+w-c}\right].
\end{equation}
As a result, the filter $\mathbf{k}_l$ with different initialization is applied to generate the feature map $\mathbf{z}_{l,j}$ for the moving window $\mathbf{W}_j$. Then, maximum pooling function\cite{boureau2010theoretical} $\operatorname{pool()}$ is used to obtain the $l$-th temporal feature, $\mathbf{y}_{l,j}=\operatorname{pool}(\mathbf{z}_{l,j}, p, s)$ by reducing the size of the feature map according to the maximum value from the feature map $\mathbf{z}_{l,j}$, where $p$ is the pooling size, and $s$ is the step size. With the FC layer after the 1D CNN, the sequence of the abstract feature $\mathbf{H}_j$ is obtained from the temporal features for the moving window $\mathbf{W}_j$. 
The sequence of the abstract feature $\mathbf{H}_j$ corresponding to the $j$-th moving window $\mathbf{W}_j$ is expressed as,
$\mathbf{H}_j=\left[\mathbf{h}_{1} ; \mathbf{h}_{2} ; \cdots ; \mathbf{h}_{i}; \cdots; \mathbf{h}_{w}\right]$, where $\mathbf{h}_{i}$ represents the abstract feature with dimension of  $1 \times m$.

\subsection{Attention Mechanism and Degradation Attention Weights}
Inspired by the success and breakthrough of the transformer model in NLP tasks\cite{vaswani2017attention}, we introduce an attention mechanism to calculate the degradation attention weights relevant to the current health states.
The degradation attention weights can effectively reflect the degeneration development.
The attention mechanism generates degradation attention weights according to the similarities between abstract features and the health states.
The initial states greatly affect the degradation development in a physical system. In order to obtain the degradation attention weights from the sequence of abstract features $\mathbf{H}_j$, the reference sequence $\mathbf{H}_r$ in terms of the initial abstract feature $\mathbf{h}_1$ of the moving window $\mathbf{W}_j$ is defined as $[\mathbf{h}_1;\cdots \mathbf{h}_1;\cdots; \mathbf{h}_1]$ with the same length of $\mathbf{H}_j$.
Other two sequences $\mathbf{H}_s$ and $\mathbf{H}_m$ in terms of the difference and product of $\mathbf{H}_j$ and $\mathbf{H}_r$, are defiend as, 
$\mathbf{H}_{s} = [0 ; \mathbf{h}_2-\mathbf{h}_1 ; \cdots; \mathbf{h}_i -\mathbf{h}_1;  \cdots;  \mathbf{h}_{n}-\mathbf{h}_1]$ and $\mathbf{H}_{m}=[\mathbf{h}_1 \cdot \mathbf{h}_1; \mathbf{h}_{2} \cdot \mathbf{h}_1 ; \mathbf{h}_i\cdot \mathbf{h}_1; \cdots ; \mathbf{h}_{n}\cdot \mathbf{h}_1 ]$.
Then, the four matrices $\mathbf{H}_j$,  $\mathbf{H}_r$,  $\mathbf{H}_s$ and  $\mathbf{H}_m$, are stacked together
$\mathcal{H}_j=[\mathbf{H}_j \; \mathbf{H}_r \; \mathbf{H}_s \; \mathbf{H}_m]$
for the attention layer. The attention layer is composed of one MLP layer and softmax function. $\mathcal{H}_j$ is fed into the MLP layer to generate the hidden state $\mathbf{u}_{i}$ as,
\begin{equation}
	\mathbf{u}_{i}=\operatorname{tanh}\left(\mathbf{W} \mathbf{h}_{i}+\mathbf{b}\right)
\end{equation}
where $\mathbf{W}$ and $\mathbf{b}$ are weight and the bias of MLP, respectively.
Afterwards, a softmax function is employed to calculate the attention weight according to correlation score between $\mathbf{u}_{i}$ and a randomly initialized vector $\mathbf{u}_{s}$. The higher score is, the stronger correlation is, and the abstract feature should be assigned with higher degradation attention weight.
After that, the degradation attention weight $\lambda_{i}$ is computed with softmax function as,
\begin{equation}
	\lambda_{i}=\frac{\exp \left(\mathbf{u}_{i}^{T} \mathbf{u}_{s}\right)}{\sum\limits_{i} \exp \left(\mathbf{u}_{i}^{T} \mathbf{u}_{s}\right)}
\end{equation}

\begin{table}[!t]
	\renewcommand\arraystretch{2}
	\centering
	\caption{Layer details of the proposed TDDN model}
	\label{layerdetails}
	\begin{tabular}{m{2cm}<{\centering}m{2cm}<{\centering}m{3cm}<{\centering}m{4.5cm}<{\centering}}
		\hline
		TDDN                                  & Layer number & Description                   & Details                     \\
		\hline
		\makecell[c]{1D CNN}                                  & Layer 1       & 1D-convolution 1D-max-pooling &
		\makecell[c]{$L$=32, filter\_size=$2\ast m$, \\
			stride=1, activation=ReLU, \\
			$p$=2, $s$=2}
		\\
		& Layer 2       & 1D-convolution 1D-max-pooling &
		\makecell[c]{$L$=64, filter\_size=64, \\
			stride=1, activation=ReLU, \\
			$p$=2, $s$=2}                                                                                        \\
		& Layer 3       & 1D-convolution 1D-max-pooling & 
		\makecell[c]{$L$=128, filter\_size=128, \\
			stride=1, activation=ReLU, \\
			$p$=2, $s$=2}                                                                                        \\                                                     
		& Layer 4       & Fully connection               & \makecell[c]{layer\_size=$w\ast m$\\
			activation=ReLU    }                                                                                                \\
		\hline
		\makecell[c]{Attention\\Mechanism}
		& Layer 5       & MLP              &
		\makecell[c]{layer\_size=$w$ \\
			activation=Tanh}
		\\
		& Layer 6       & \makecell[c]{Attention-weighted\\ state }                   & softmax function
		\\
		\hline
		\makecell[c]{Regression} & Layer 7       & Fully connection               & \makecell[c]{layer\_size=8 \\
			activation=ReLU                                }                                                                    \\
		& \makecell[c]{Layer 8}       & Regression                    & layer\_size=1               \\
		\hline
		\makecell[c]{Hyper-\\parameters} & Learning rate		& Batch size		& Epochs
		\\
		& \makecell[c]{$10^{-4}$ ($10^{-5}$)}		& 32		&	200
		\\
		& \makecell[c]{Loss}	& Optimizer	& Window size
		\\
		& \makecell[c]{Mean square\\ error}	& Adam	& 64
		\\
		\hline
	\end{tabular}
\end{table}
With the attention weights, TDDN can identify critical degradation patterns, so as to capture all the relevant abstract features. The attention-weighted state $\mathbf{s}_j$ is a weighed sum of all extracted abstract features for the moving window $W_j$,
\begin{equation}
	\mathbf{s}_j=\sum_{i} \lambda_{i} \mathbf{h}_{i}
\end{equation}
Finally, the attention-weighted state $\mathbf{s}_j$ is fed into the fully connected layer to make the RUL prediction at the $j$-th time step.
In summary, the structure parameters of TDDN is given in Table ~\ref{layerdetails}

\section{Experimental Study: A Small Fleet of Turbofan Engines}\label{results}

This section describes the process of data preprocessing, sample construction, and label setting in the Commercial Modular Aero-Propulsion System Simulation (C-MAPSS) dataset. Then, the proposed method is evaluated on a benchmark dataset and compared with the latest prediction results.

\subsection{Data Preprocessing}\label{data}
The C-MAPSS dataset provided by NASA is widely used to evaluate the performance of models. It is generated by a thermo-dynamical simulation model which simulates damage propagation and performance degradation in MATLAB and Simulink environment\cite{saxena2008damage,saxena2008turbofan}. In addition, considering noisy fluctuations in real data, random measurement noises are added to the sensor output to enhance the authenticity of data.
According to engine operating conditions and fault modes, the dataset is divided into four subsets. In each subset, engine number, operational cycle number, three operational settings, and 21 sensor measurements reflect turbofan engine degradation.
Since each engine normally works at the start and then occurs fault at a random time, the degradation process is generally different from one to another. The detailed description of sensor data can be found in the previous work\cite{saxena2008damage}.
Each subset is further divided into a training dataset and a test dataset. Finally, the operating conditions, fault modes, and other statistical data of each subset are listed in Table~\ref{C-MAPSS}.
Especially, the training dataset records the whole run-to-failure data. In contrast, the test dataset terminates some time prior to the machinery failure. The corresponding RUL of each engine is included in the dataset for verification purposes. Therefore, the objective is to predict the RUL of each engine based on provided sensor streaming data in the test dataset.

\begin{figure}[!h]
	\centering
	\includegraphics[width=8cm]{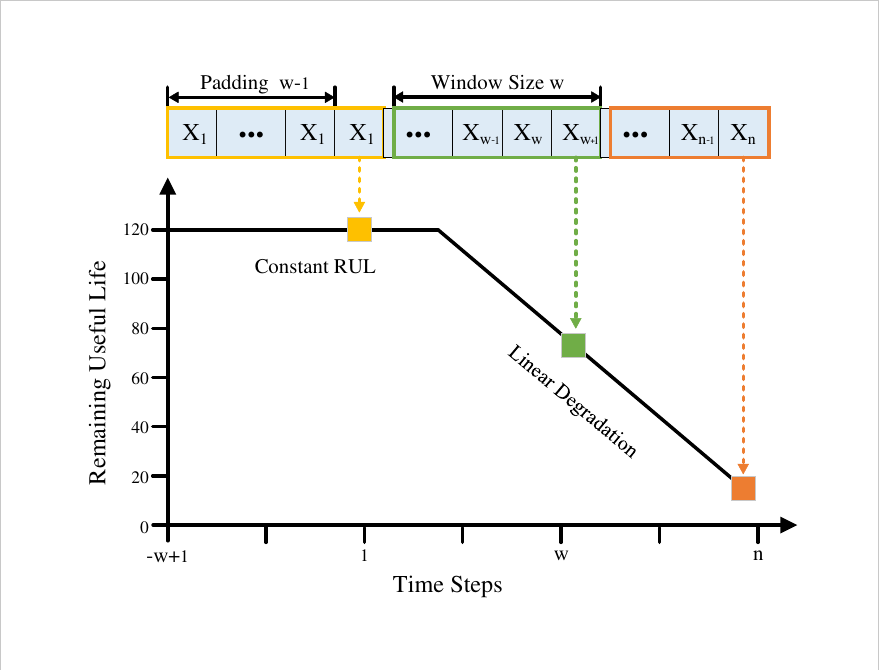}
	\caption{Illustration of data segmentation using moving windows to generate input samples}
	\label{timewindow}
\end{figure}

The C-MAPSS dataset only provides the RUL corresponding to the last cycle of each engine in the test dataset. The RUL labels are added to the training dataset to establish the relationship between the sensor data of the turbofan engine and RUL.
In practical applications, the degradation process of turbofan engines is very long, and the degradation trend in the early and healthy operation stages is not distinct, so the degradation of the system is usually negligible.
When a fault occurs at a certain time point, the engine performance will be reduced, and when the engine reaches complete fault, the engine life will be terminated. Therefore, this paper uses a piecewise linear function which is shown in Fig.~\ref{timewindow} to set the RUL label of the training dataset. Besides, it is observed that the degradation usually occurred at the late 120-130 cycles\cite{heimes2008recurrent}. Therefore, the maximum value of RUL is set to be 120 based on numerous experiments.

\begin{table}
	\renewcommand\arraystretch{2}
	\centering
	\caption{Information of the C-MAPSS dataset}
	\begin{tabular}{m{4cm}<{\centering}m{2cm}<{\centering}m{2cm}<{\centering}m{2cm}<{\centering}m{2cm}<{\centering}}
		\hline
		Dataset                                & FD001 & FD002 & FD003 & FD004 \\
		\hline
		Engines in training data               & 100   & 260   & 100   & 259   \\

		Engines in test data                & 100   & 259   & 100   & 248   \\

		Operating conditions                   & 1     & 6     & 1     & 6     \\

		Fault modes                            & 1     & 1     & 2     & 2     \\

		\makecell[c]{Minimum running cycle\\ in training data} & 128   & 128   & 145   & 128   \\

		\makecell[c]{Minimum running cycle\\ in test data}  & 31    & 21    & 38    & 19    \\

		\hline
	\end{tabular}
	\label{C-MAPSS}
\end{table}

Sensor signals in real system has fluctuations in magnitudes and units, so not all raw  sensor data are useful to the RUL prediction. Each sensor streaming data is normalized to be within the range of $\left[-1, 1\right]$ by using min-max scaler. To formulate the scaling function, let the multivariate time series $\mathbf{X}=\left[\mathbf{x}_{1} \; \mathbf{x}_{2} \; \cdots \; \mathbf{x}_{j} \; \cdots \; \mathbf{x}_{m}\right]$, where $\mathbf{x}_{j}=\left[x_{1,j}\; {x_{2,j} \; \cdots \;  x_{n,j}}\right]^T$ represents the time series data of $i$-th sensor. The $i$-th normalized value is calculated as,
\begin{equation}
	\mathbf{x}_{j}=\frac{2\left(\mathbf{x}_{j}-\min \left(\mathbf{x}_{j}\right)\right)}{\max \left(\mathbf{x}_{j}\right)-\min \left(\mathbf{x}_{j}\right)}-1
\end{equation}
where,  $\min \left(\mathbf{x}_{j}\right)$ and $\max \left(\mathbf{x}_{j}\right)$ denote the maximum and minimum values in the vector $\mathbf{x}_{j}$, respectively. Moreover, it is noted that the normalization of the test dataset is based on the maximum and minimum values of the training dataset.

\begin{figure}[!h]
	\centering
	\includegraphics[width=8cm]{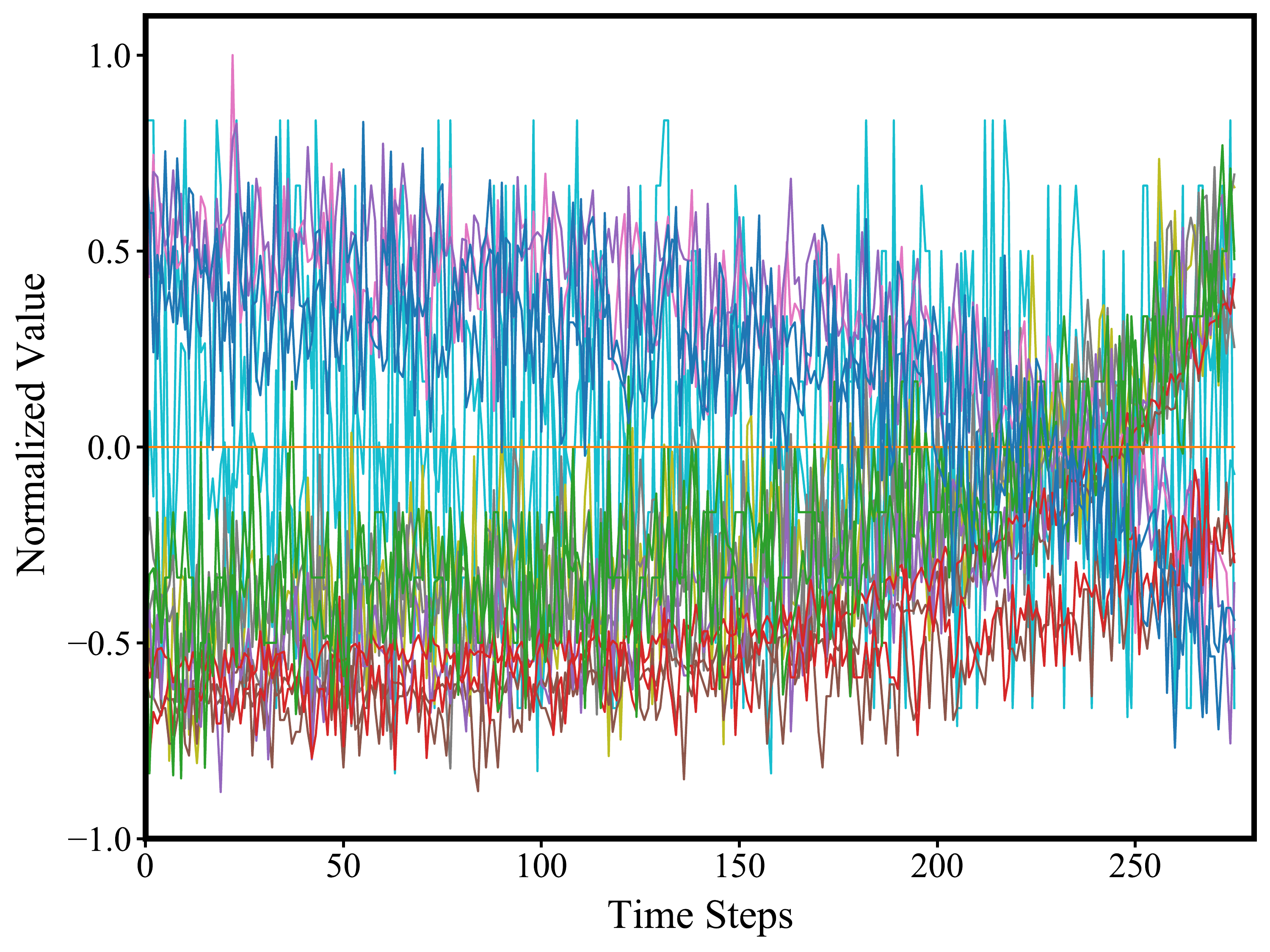}
	\caption{The visualization of sensor streaming data and settings in FD001}
	\label{sensorFD001}
\end{figure}

\begin{figure}[!h]
	\centering
	\includegraphics[width=8cm]{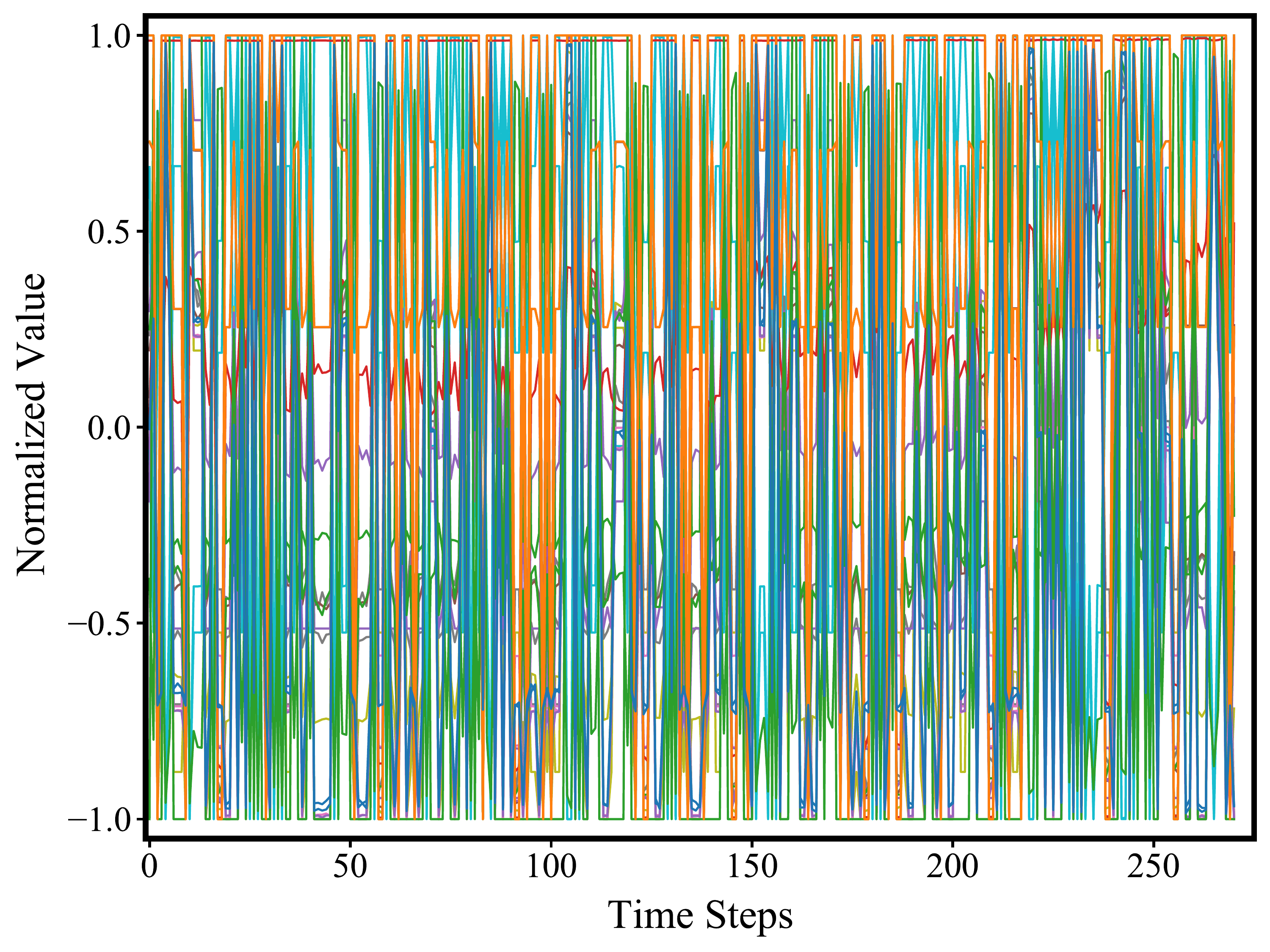}
	\caption{The visualization of sensor streaming data and settings in FD004}
	\label{sensorFD004}
\end{figure}

The selection is carried out to remove irrelevant sensor streaming data to improve computational efficiency and save training time. FD001 and FD003 datasets have a single operating condition.
In the FD001 and FD003 datasets, there are twenty-one sensors labeled with the indices of $(1,\cdots, 21)$ and three operational settings of s1,s2, and s3.
In order to select the relevant sensor streaming data, the twenty-one sensors and operational settings in the FD001 dataset are visualized in Fig.~\ref{sensorFD001}. The sensor measurements and operational settings demonstrate the ascending, descending, and unchanged three trends throughout the whole life in the FD001 dataset.
According to the ascending, descending, and unchanged trends, we sort the sensor and operation settings into three trend categories listed in Table~\ref{sensortrend}.
It is noticed that most sensors have clearly ascending or descending trends in the degradation trajectories, while s3 and sensors 1, 5, 6, 10, 16, 18, and 19 have the unchanged trend that provides no useful degradation information for the RUL prediction.
Hence, the streaming data in sensors and settings with the ascending and descending trends in Table~\ref{sensortrend} are used as the inputs for the TDDN model.
Due to the same operating condition and degradation pattern, the same trend categories also exist in the FD003 dataset.
However, selecting sensors in the FD002 and FD004 datasets is more challenging because they have six operating conditions.
Since sensors and settings in the FD004 dataset have fluctuating trajectories in Fig.~\ref{sensorFD004}, it is impossible to select the sensors in the visualization.
Therefore, all operational settings and sensor measurements are used for the RUL prediction of the FD002 and FD004 datasets.

\begin{table}[!h]
	\renewcommand\arraystretch{2}
	\rmfamily
	\centering
	\caption{\rmfamily The trend categories of the sensors and operational settings in FD001.}
	\begin{tabular}{m{2cm}<{\centering}|m{4cm}<{\centering}}
		\hline
		Trend      & Sensors                       \\
		\hline
		Ascending  & 2,3,4,8,9,11,13,15,17, s1, s2 \\
		
		Descending & 7,12,20,21, s1, s2            \\
		
		Unchanged  & 1,5,6,10,14,16,18,19, s3      \\
		
		\hline
	\end{tabular}
	\label{sensortrend}
\end{table}

In addition, it should be noted that the sensor streaming data in different engines have different lengths,
as shown in Table~\ref{C-MAPSS}. Cycle length discrepancy limits the maximum size of the moving window and affects the RUL prediction accuracy.
The smaller size of moving window may result in the larger prediction error.
In order to eliminate the size limitation of the moving window, we pad the sensor streaming time with the first cycle value $\mathbf{x}_1$ $w-1$ times as,
\begin{equation}
	\mathbf{X}=[\underbrace{\mathbf{x}_{1} \; \cdots \; \mathbf{x}_{1}}_{w} \; \mathbf{x}_{2} \; \cdots \; \mathbf{x}_{n-1} \; \mathbf{x}_{n}]
\end{equation}
Moving windows and padding strategy are illustrated in Fig.~\ref{timewindow}.
The padding strategy allows utilizing the engines with fewer cycles rather than removing them from samples in the traditional approach\cite{babu2016deep,li2018remaining}. With the padding, the engine RUL can be predicted even if only a small number of cycles are given.

\subsection{TDDN Training}

\begin{figure}[!h]
	\centering
	\includegraphics[width=8cm]{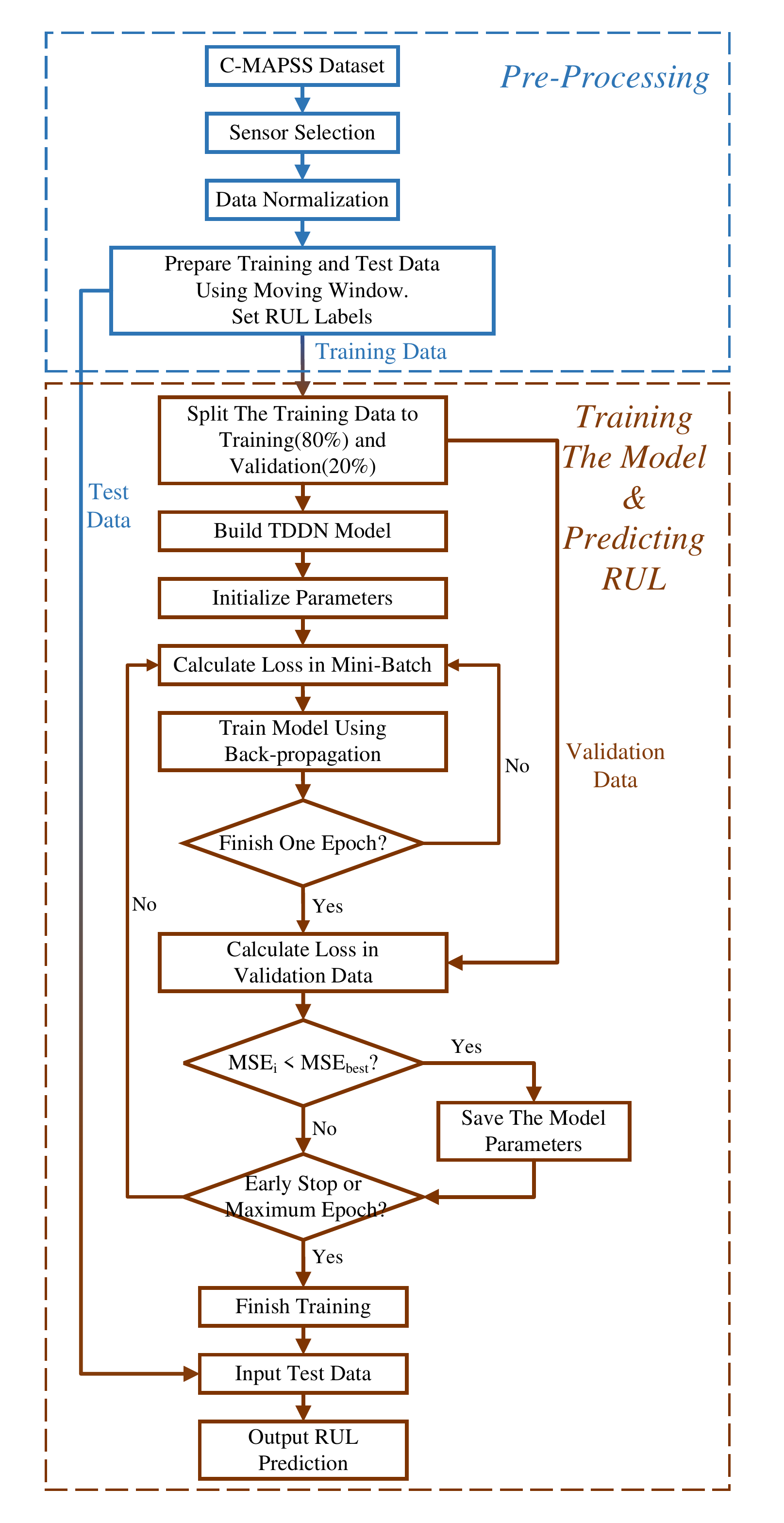}
	\caption{Flowchart of TDDN model for prognostics}
	\label{flowchart}
\end{figure}

Fig.~\ref{flowchart} illustrates the complete procedure of the TDDN model to predict the turbofan engine RUL.
Data visualization can help select meaningful sensor streaming data since some data remain constant and contain no useful information.
All the selected sensor data are normalized.
Significantly, sensor streaming data are padded with the first cycle values in the beginning part to improve window size. The training data is fed into the TDDN prediction method which effectively extracts the temporal features and captures key fault characteristics. All the TDDN model hyperparameters are shown in Table~\ref{layerdetails}.

The validation dataset is used to evaluate the performance corresponding to each training epoch to verify the effectiveness of the TDDN model and select appropriate hyperparameters.
Hence, 20\% engines in each training dataset are randomly selected as validation sets while the rest engines are training samples.
Through experiments, the batch size of 32 and the maximum epoch of 200 can achieve the best prediction performance.
Meanwhile, the learning rate is set to be 0.0001 initially for the fast optimization at the 100 epochs and then reduces to one-tenth of the initial learning rate for stable convergence. The back-propagation learning is adopted to update the weights in the network to minimize the mean square error (MSE) loss function. At the same time, the Adam optimizer is used as the network gradient descent algorithm. In addition, an early stopping strategy is applied to mitigate the overfitting problem. The training process terminates in advance when the validation performance shows no improvement of more than ten consecutive epochs.

\subsection{Performance Benchmark}

Two performance metrics, namely root mean square error (RMSE) and scoring function, are generally used to establish a common comparison with state-of-the-art methods. They are defined as,
\begin{equation}
	{\rm RMSE}=\sqrt{\frac{1}{n} \sum_{i=1}^{n} d_{i}^{2}}
\end{equation}

\begin{equation}
	{s}=\left\{
	\begin{aligned}
		 & \sum_{i=1}^{n}\left(e^{-\frac{h}{13}}-1\right), & d_{i}<0      \\
		 & \sum_{i=1}^{n}\left(e^{\frac{h}{10}}-1\right),  & d_{i} \geq 0
	\end{aligned}\right.
\end{equation}

\begin{figure}[!h]
	\centering
	\includegraphics[width=8cm]{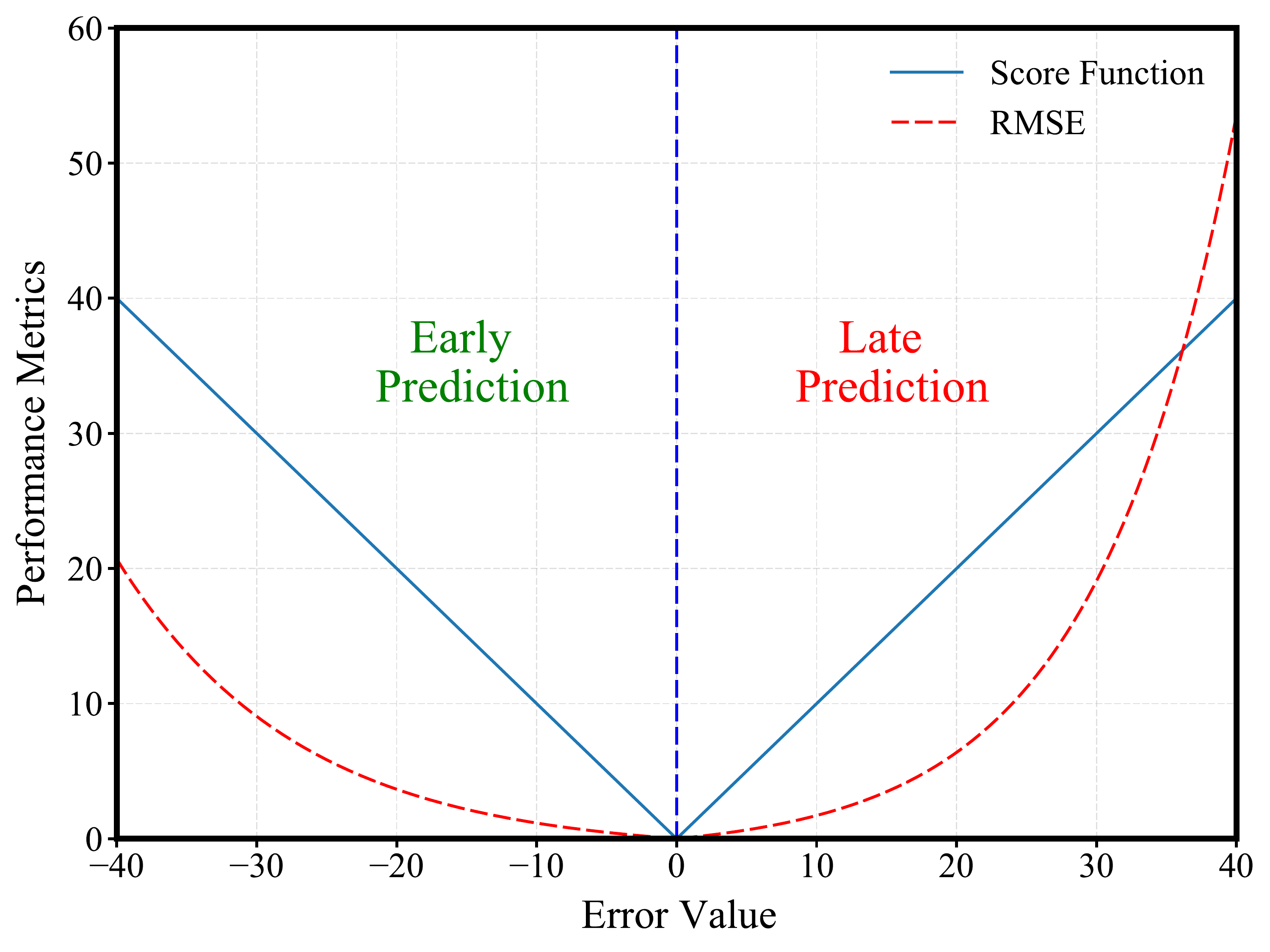}
	\caption{Comparison between scoring function and RMSE for different error values}
	\label{functioncompare}
\end{figure}

Where $s$ denotes the score, and $n$ represents the total number of samples in the test dataset. $d_{i}=RUL^{'}_{i}-RUL_{i}$, is the difference between the $i$-th sample predicted RUL value and true RUL value.
The comparison between the two performance metrics is shown in Fig.~\ref{functioncompare}.
The asymmetric scoring function penalizes the late prediction error more than the early prediction error, while RMSE gives an equal penalty to prediction error.
The late prediction error usually leads to severe consequences for a mechanical degradation scenario, while early prediction error can allow maintenance in advance and avoid unnecessary downtime.
However, a single outlier significantly affects this scoring function value since it is exponential and has no error normalization. Accordingly, it is noted that a single scoring function cannot thoroughly evaluate the performance of an algorithm. Therefore, performance metrics are a combined aggregate of RMSE and scoring function that ensures an algorithm consistently predicts accurately and timely.

The performance of the TDDN model is compared with the existing methods on all subsets in the C-MAPSS dataset according to the performance metrics of RMSE and Score listed in Table~\ref{modelcompare}.
The results indicate that the proposed method achieves noticeable improvement and has a superior performance in different scenarios.
Specifically, despite the score of the proposed method is similar to the best score in previous literature on FD001 and FD003 datasets, the proposed method obtains lower RUL standard deviations in the degradation process.
Observing the results of FD002 and FD004 datasets, the proposed method remarkably outperforms other methods with the lowest prediction errors.
Significantly, compared with the best previous results, it decreased by more than 28\% for RMSE and 56\% for the score on the FD002 dataset, and 38\% for RMSE and 35\% for the score on the FD004 dataset.
In this way, the precision and reliability of the proposed TDDN model in real-life PHM applications have been successfully validated.
Besides, all existing methods have higher prediction errors on FD002 and FD004 datasets than that on two other datasets because of more complex operating conditions.
Thanks to extracting degradation-related features, the lowest difference in prediction error between 4 subsets are achieved by the TDDN model.
Furthermore, the TDDN model in complex conditions is also able to achieve the optimal prediction accuracy of other methods in normal conditions. The substantial performance improvements illustrate the superiority of the TDDN model in complex conditions.
It is also worth noting that most industrial machinery work in a variety of complex conditions in the real world, so the proposed method is quite suitable for the practical applications of machinery.

\begin{table*}[!h]
	\rmfamily
	\centering
	\caption{The benchmarking of the TDDN model on the C-MAPSS dataset}
	\renewcommand\arraystretch{1.5}{
	\begin{tabular}{m{1.5cm}<{\centering}m{1cm}<{\centering}m{1cm}<{\centering}m{1cm}<{\centering}m{1cm}<{\centering}|m{1cm}<{\centering}m{1cm}<{\centering}m{1cm}<{\centering}m{1cm}<{\centering}}
		\hline
		                & \multicolumn{4}{c|}{RMSE} & \multicolumn{4}{c}{Score}                                                                                                          \\
		\hline
		Method          & FD001                     & FD002                     & FD003         & FD004          & FD001           & FD002           & FD003           & FD004           \\
		\hline
		LSTM\cite{zheng2017long}            & 16.14                     & 24.49                     & 16.18         & 28.17          & 338             & 4450            & 852             & 5550            \\
		DCNN\cite{li2018remaining}            & 12.61                     & 22.36                     & 12.64         & 23.31          & 273.7           & 10412           & 284.1           & 12466           \\
		HDNN\cite{al2019multimodal}        
		& 13.02			& 15.24			& 12.22         & 18.16          & 245           & 1282.42       & 287.72      & 1527.42           \\
		MS-DCNN\cite{li2020remaining}        
		& 11.44			& 19.35			& 11.67         & 22.22          & \textbf{196.22}       & 3747          & 241.89      & 4844           \\
		Li-DAG\cite{li2019directed}          & 11.96                     & 20.34                     & 12.46         & 22.43          & 229             & 2730            & 535             & 3370            \\
		Ellefsen $\it{et\; al.}$\cite{ellefsen2019remaining}          
		& 12.56			& 22.73			& 12.10         & 22.66          & 231           & 3366          & 251           & 3370            \\
		DA-TCN\cite{song2020distributed}          & 11.78                     & 16.95                     & 11.56         & 18.23          & 229.48          & 1842.38         & 257.11          & 2317.32         \\
		Proposed TDDN model & \textbf{9.47}             & \textbf{10.93}            & \textbf{9.17} & \textbf{11.16} 	& 214.17 	& \textbf{561.82} & \textbf{216.79}          & \textbf{997.69} \\
		\hline
	\end{tabular}}
	\label{modelcompare}
\end{table*}

\begin{figure}[!t]
	\centering
	\subfigure[Test engine 34 in FD001]{
		\includegraphics[width=0.49\textwidth]{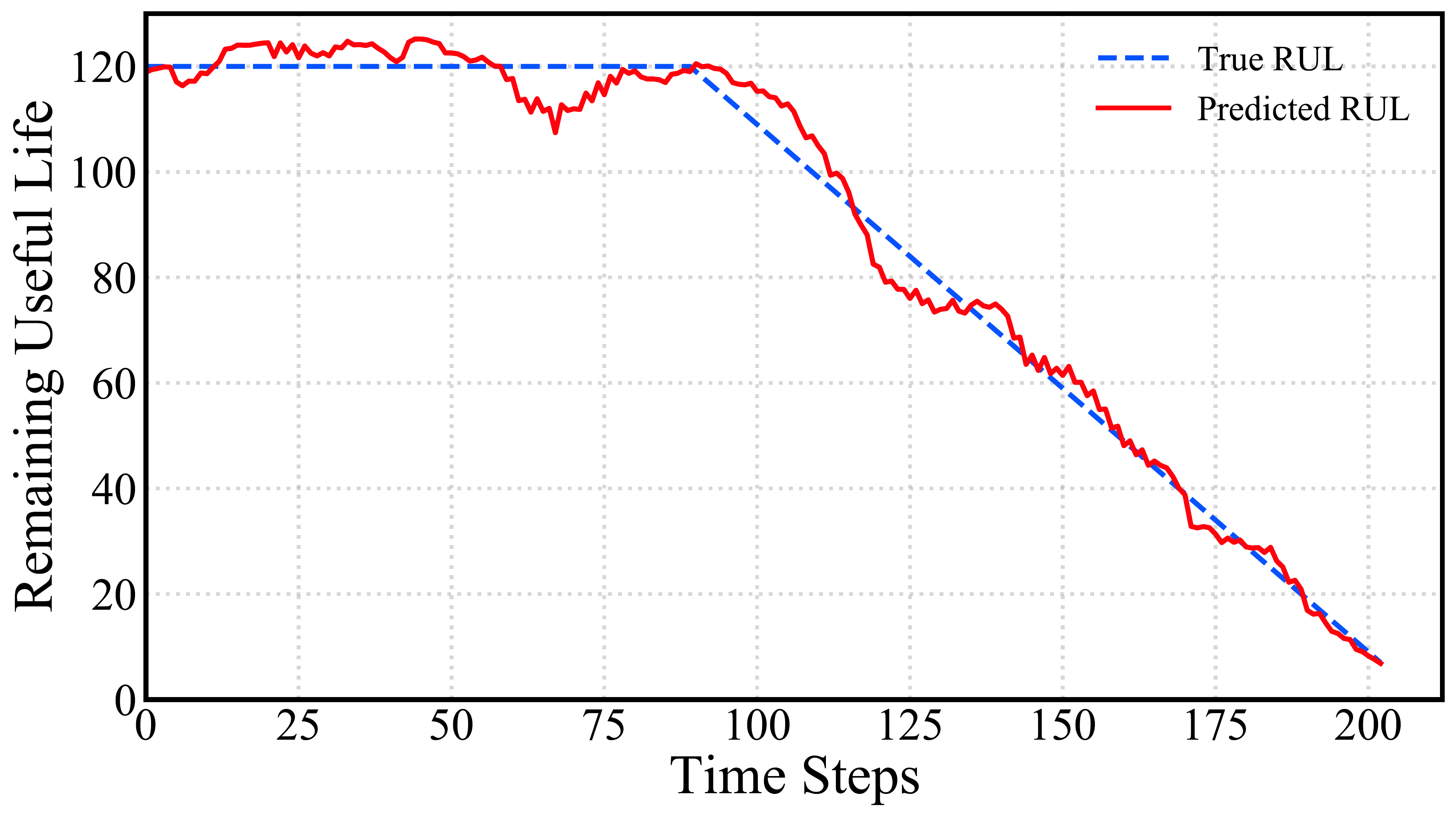}}
	\subfigure[Test engine 138 in FD002]{
		\includegraphics[width=0.49\textwidth]{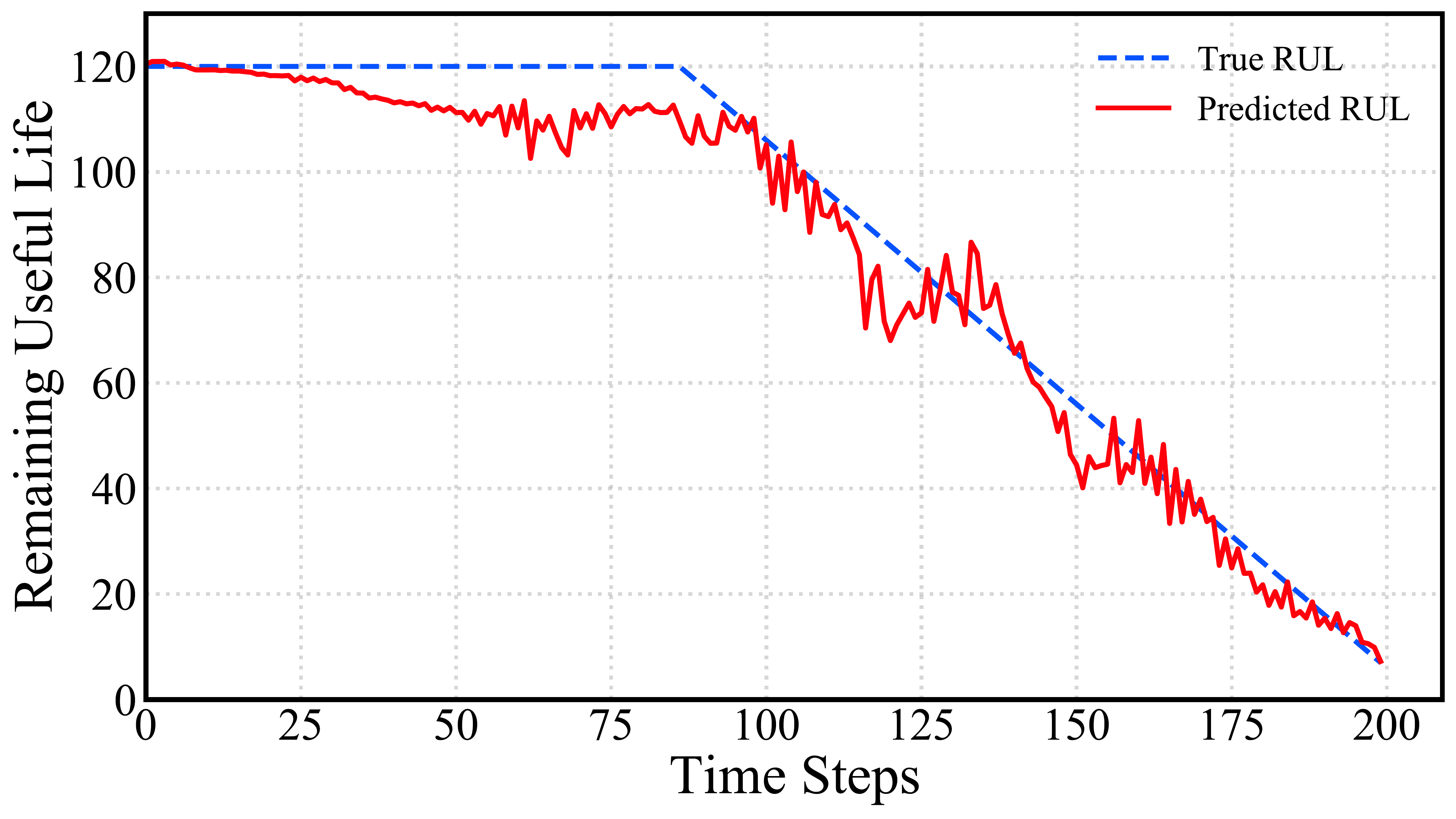}}\\
	\subfigure[Test engine 92 in FD003]{
		\includegraphics[width=0.49\textwidth]{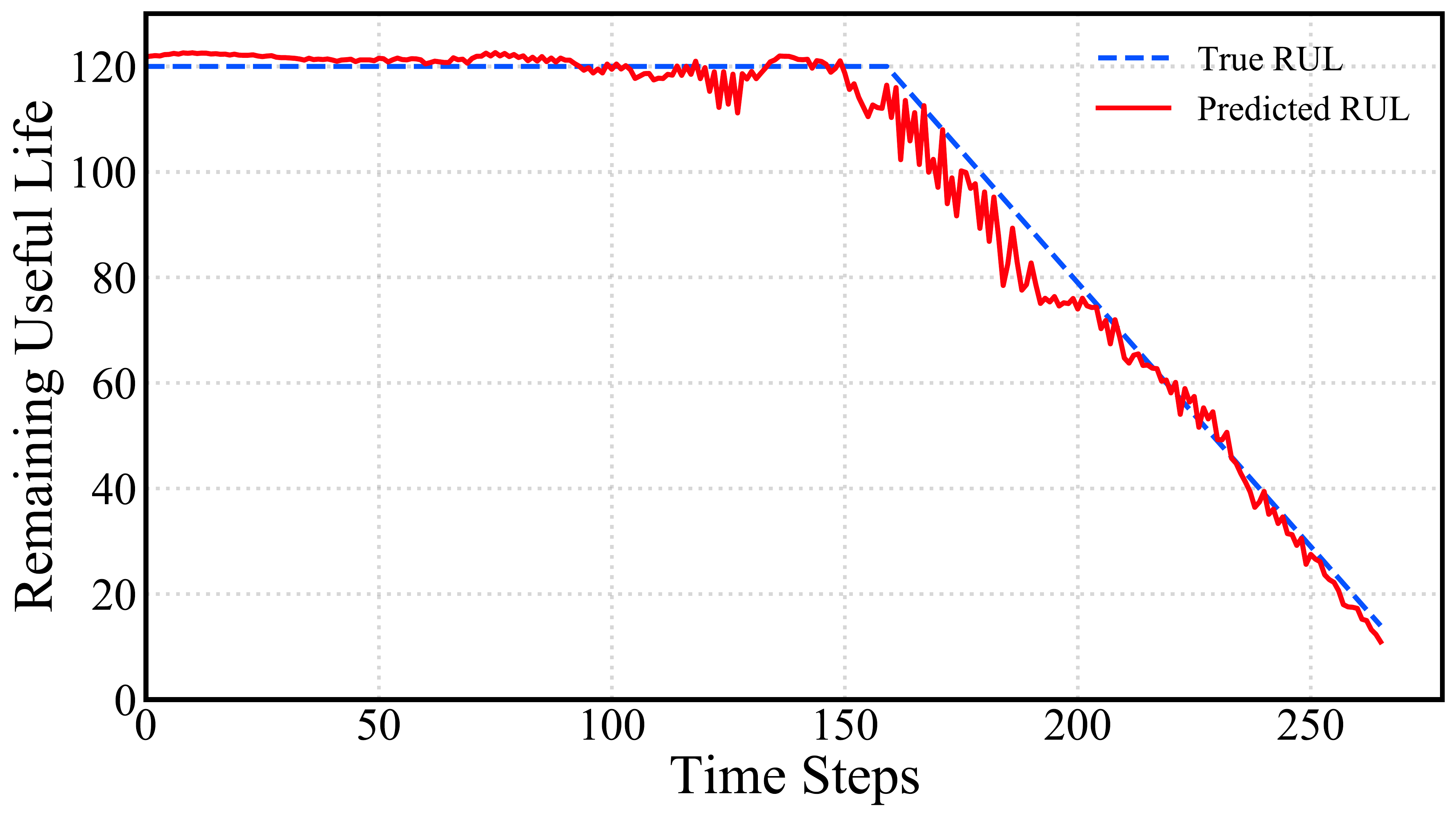}}
	\subfigure[Test engine 135 in FD004]{
		\includegraphics[width=0.49\textwidth]{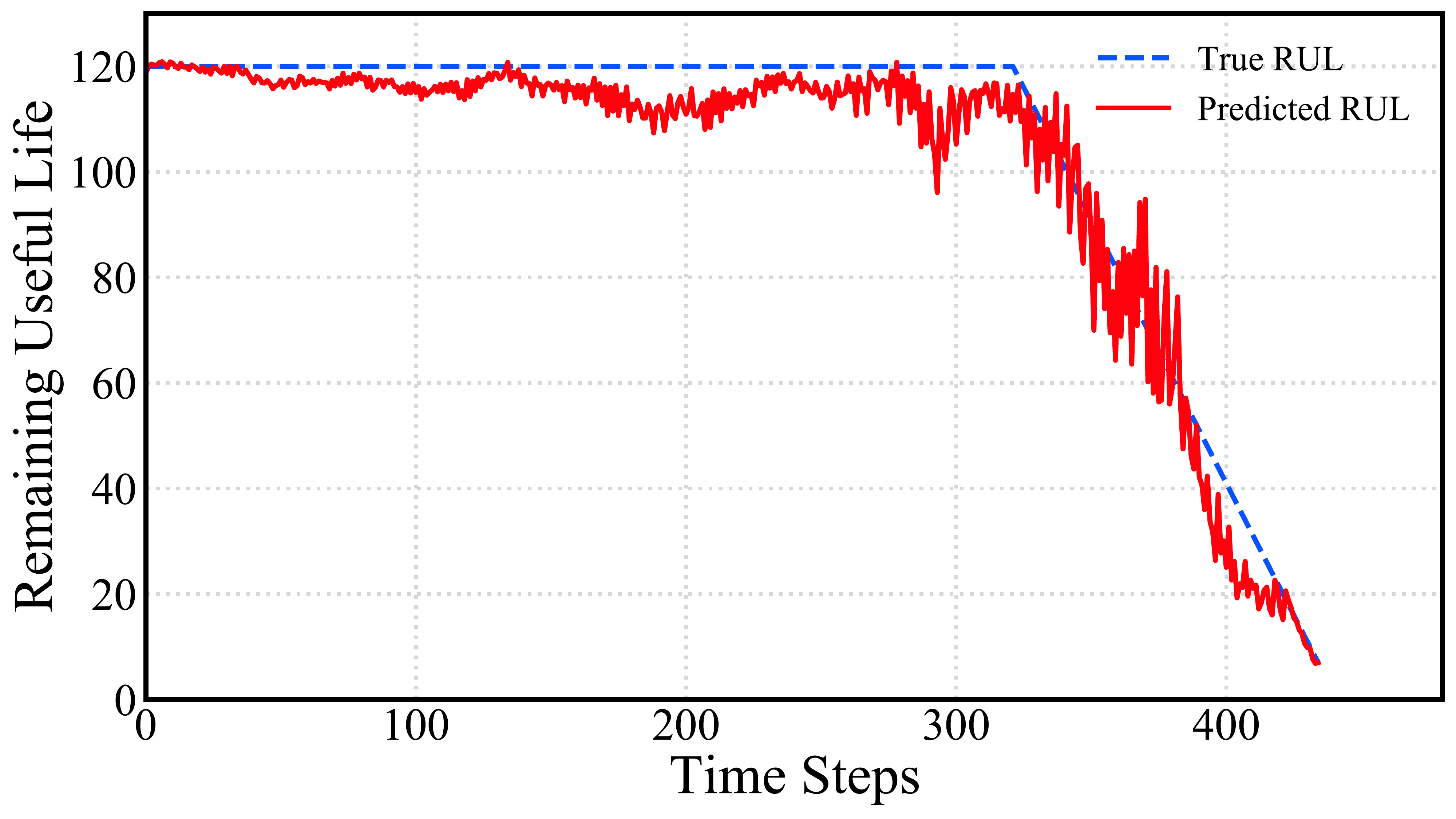}}
	\caption{RUL prediction results on different test datasets}
	\label{predict} 
\end{figure}

To visualize the prediction results, one representative engine is selected in the four datasets of FD001, FD002, FD003, and FD004.
Fig.~\ref{predict} shows that the predicted RUL evolves around the actual RUL.
TDDN can effectively capture the trend of the degradation development.
The fault characteristics accumulate at the final failure stage of the degradation development.
The prediction error gradually decreases when an engine is close to the final failure stage.
It is also observed that the predicted RUL is mostly below the actual RUL in Fig.~\ref{predict}, which reduces the chance of overestimating the RUL to cause unnecessary maintenance costs.
Hence, the proposed TDDN model can achieve outstanding prediction performance under various operating conditions in the FD001, FD002, FD003, and FD004 datasets.

\section{TDDN Hyperparameters and Performance Analysis}\label{pa}
This section studies the influence of the moving window size and the number of convolutional layers on the prediction results. In addition, the degeneration-related features extracted by 1D CNN and the attention mechanism to the TDDN model are studied.

\subsection{Size Dependence of Moving Windows and Convolutional Layers}

The sensor streaming data are segmented into moving windows. The size of the moving window is $w\times m$, where $w$ and $m$ denote the size of the moving window and the number of selected sensor data, respectively.
The label of the moving window $\mathbf{W}_j$ is determined by the RUL at the $j$-th cycle.
Moving windows take the temporal features evolution into consideration. Therefore, the tuning of moving windows can better reflect the system degradation patterns and improve the prediction performance.
Particularly, the size of the moving window plays a vital role in the prediction performance.
The small window size does not contain much useful degradation information.
\begin{figure}[!b]
	\centering
	\subfigure[FD001]{
		\includegraphics[width=0.49\textwidth]{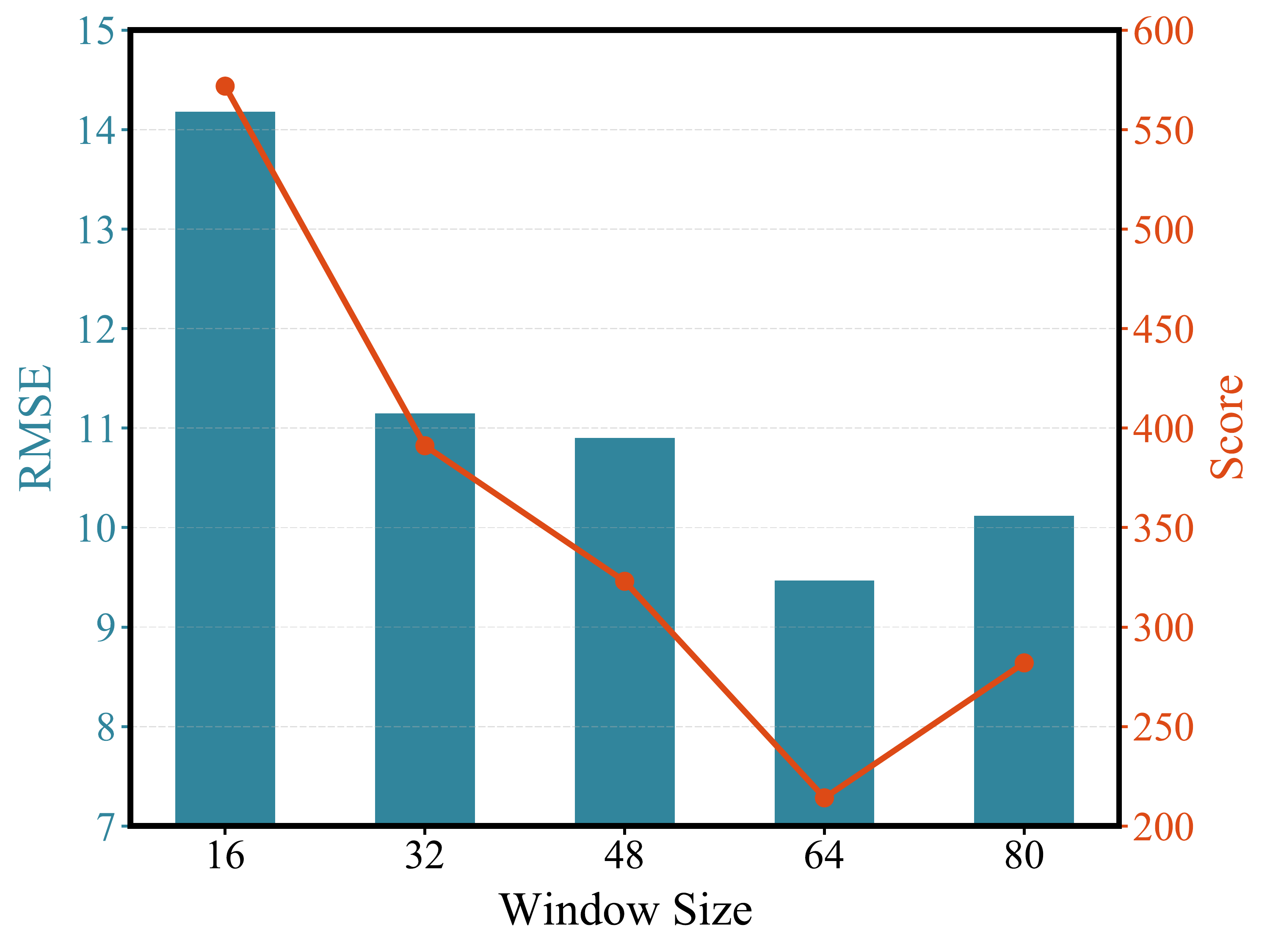}}
	\subfigure[FD002]{
		\includegraphics[width=0.49\textwidth]{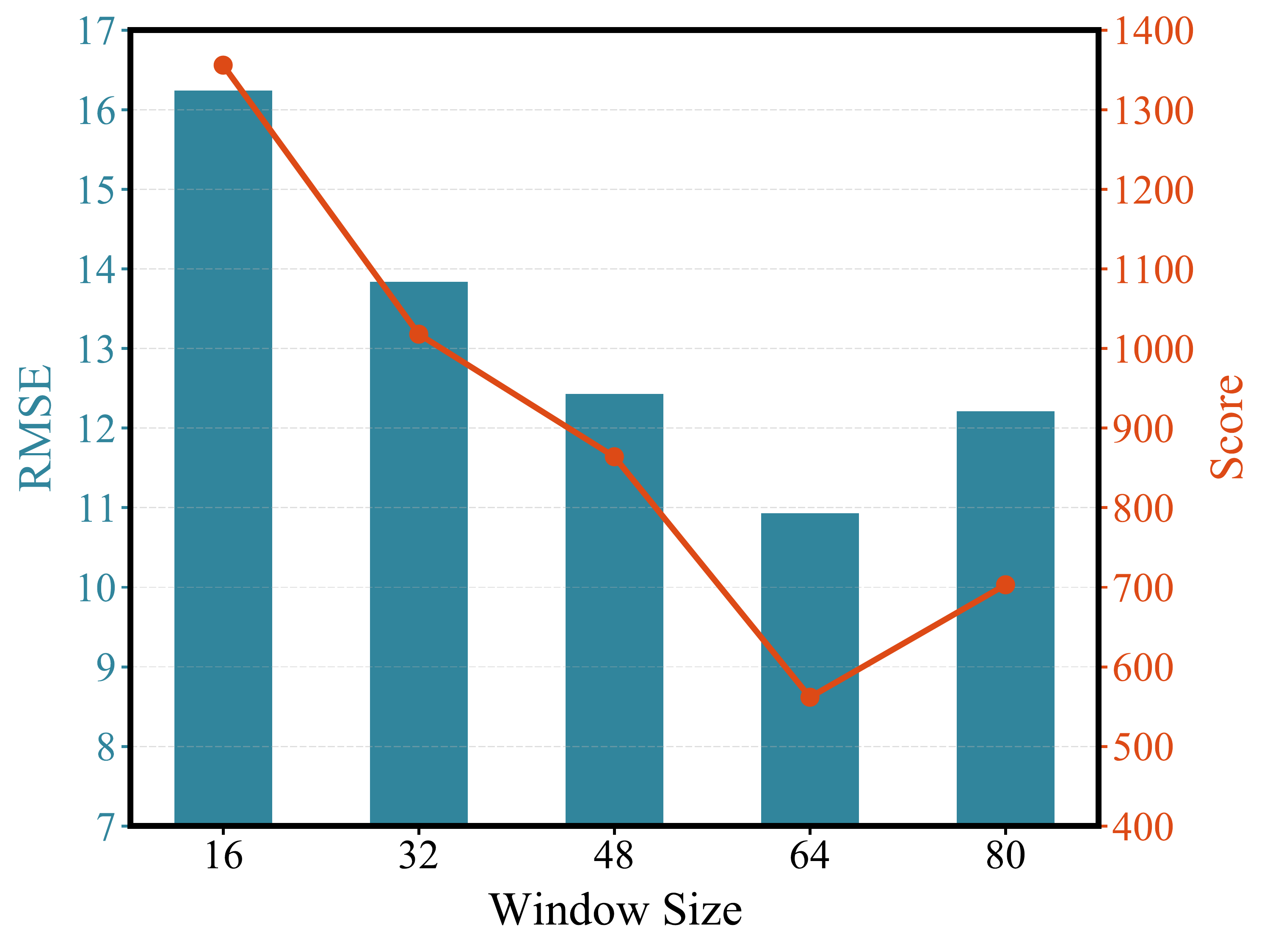}}\\
	\subfigure[FD003]{
		\includegraphics[width=0.49\textwidth]{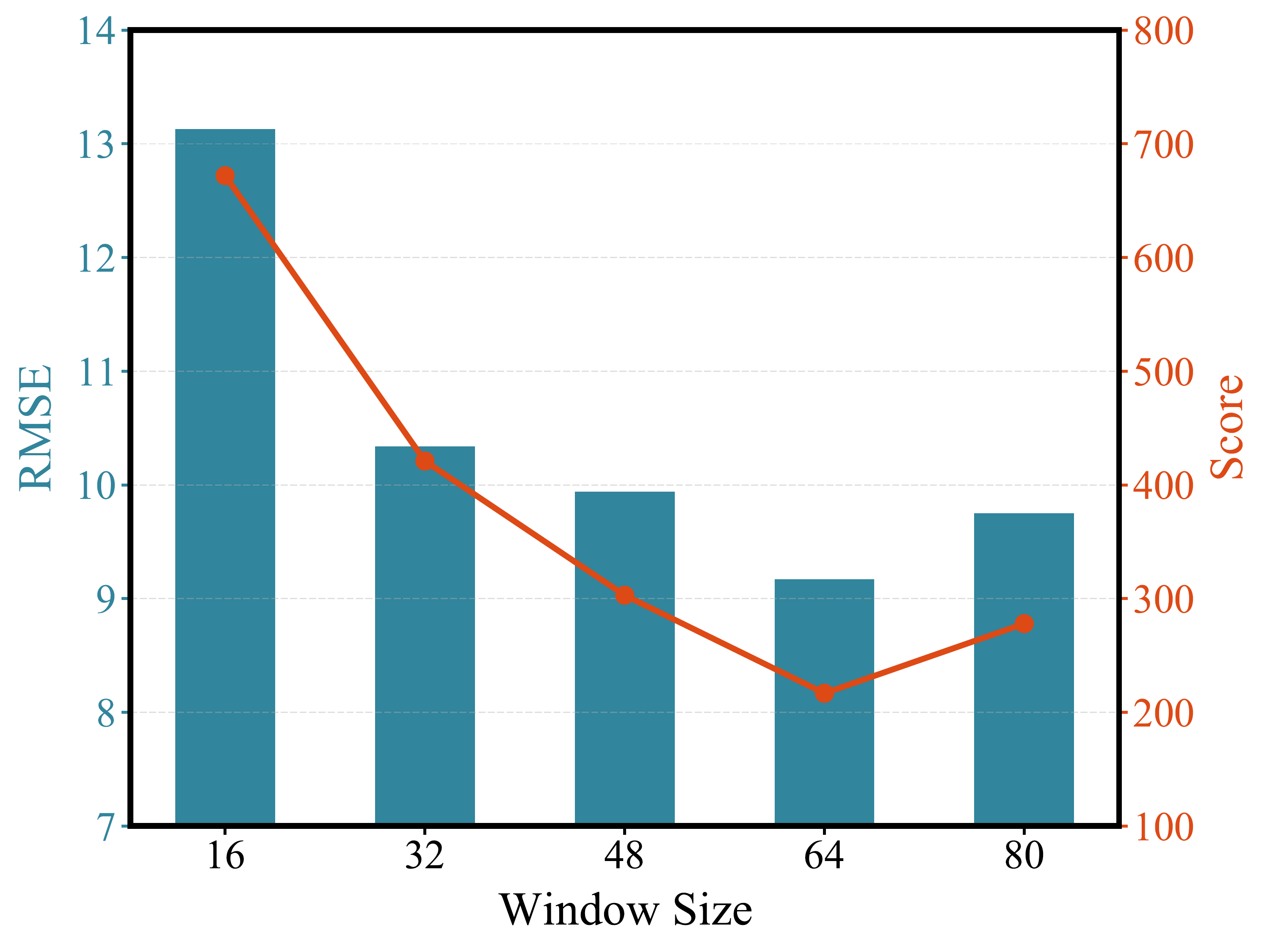}}
	\subfigure[FD004]{
		\includegraphics[width=0.49\textwidth]{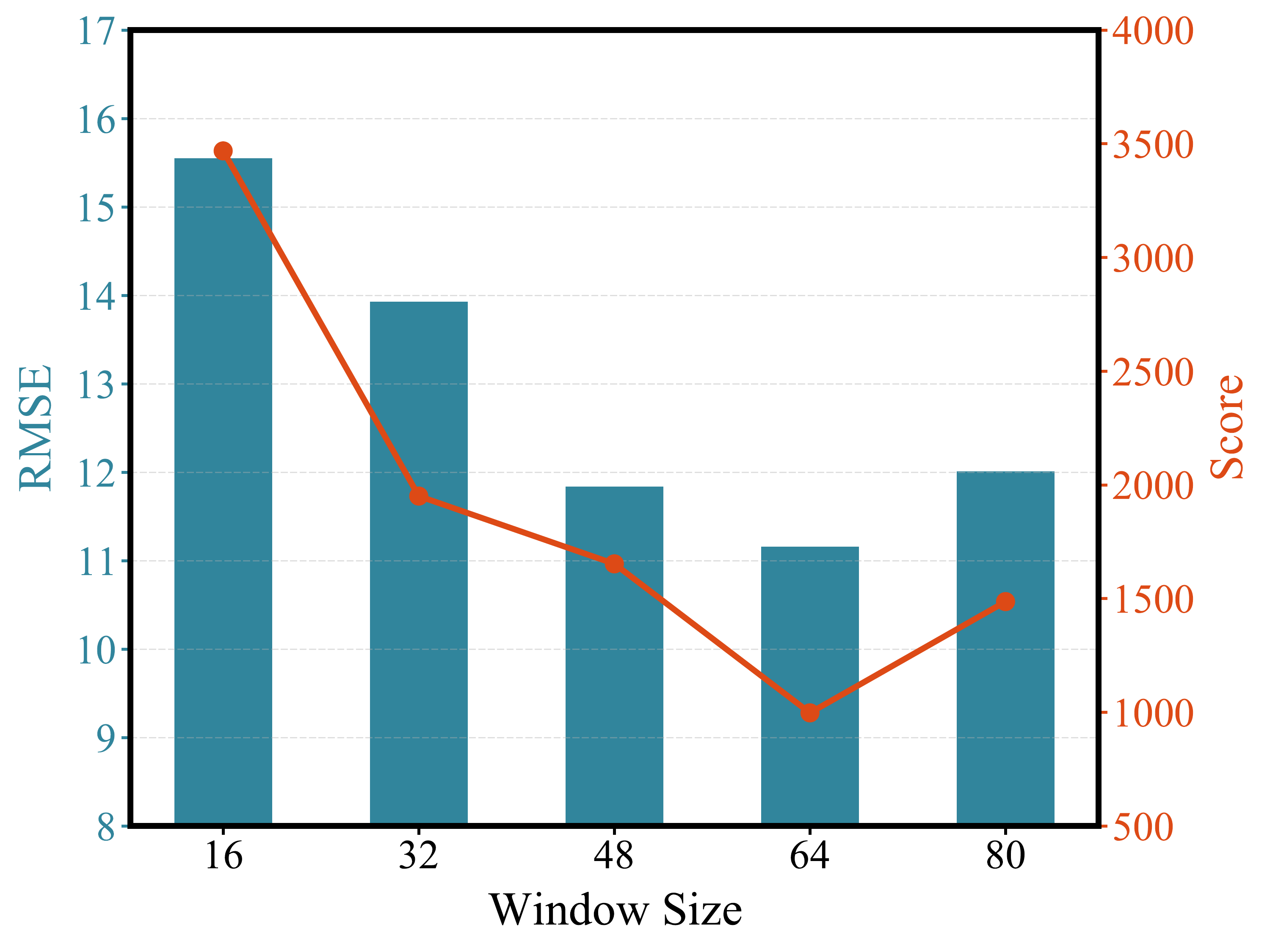}}
	\caption{The RMSE and Score performance metrics {\it{v.s.}} moving window size}
	\label{window} 
\end{figure}
On the other hand, unnecessarily large window sizes need more computational time and the loss of crucial degradation-related features.
Thus, selecting the appropriate window size is critically important to improving the prediction performance.
The minimum running cycles of engine limit the window size selection in the previous work\cite{babu2016deep,li2018remaining}.
However, with the padding strategy, the window size can be extended accordingly to make the moving window size equal for all the engines.
The window sizes ranging from 16 cycles to 80 cycles are tested on FD001, FD002, FD003, and FD004 four datasets to understand the window size dependence.
Each condition is tested five times independently, and the average value of experimental results is shown in Fig.~\ref{window}.
The temporal features and fault characteristics are contained in the optimal moving window, which could significantly enhance the extraction of degradation-related features.
The test experiments show that the larger window size can improve the prediction performance of the TDDN model within 64 cycles.
If the window size further increases from 64 cycles to 80 cycles, the TDDN model prediction performance does not change much in the FD001 and FD003 datasets but becomes poorer in the FD002 and FD004 datasets in Fig.~\ref{window}.
The unnecessarily large moving window size can diminish the performance of the attention mechanism. Based on the test experiments, the optimal moving window size is set to be 64 cycles, which can effectively balance degradation-related features extraction and computational time consumption in sensor streaming data.

\begin{figure}[!h]
	\centering
	\includegraphics[width=8cm]{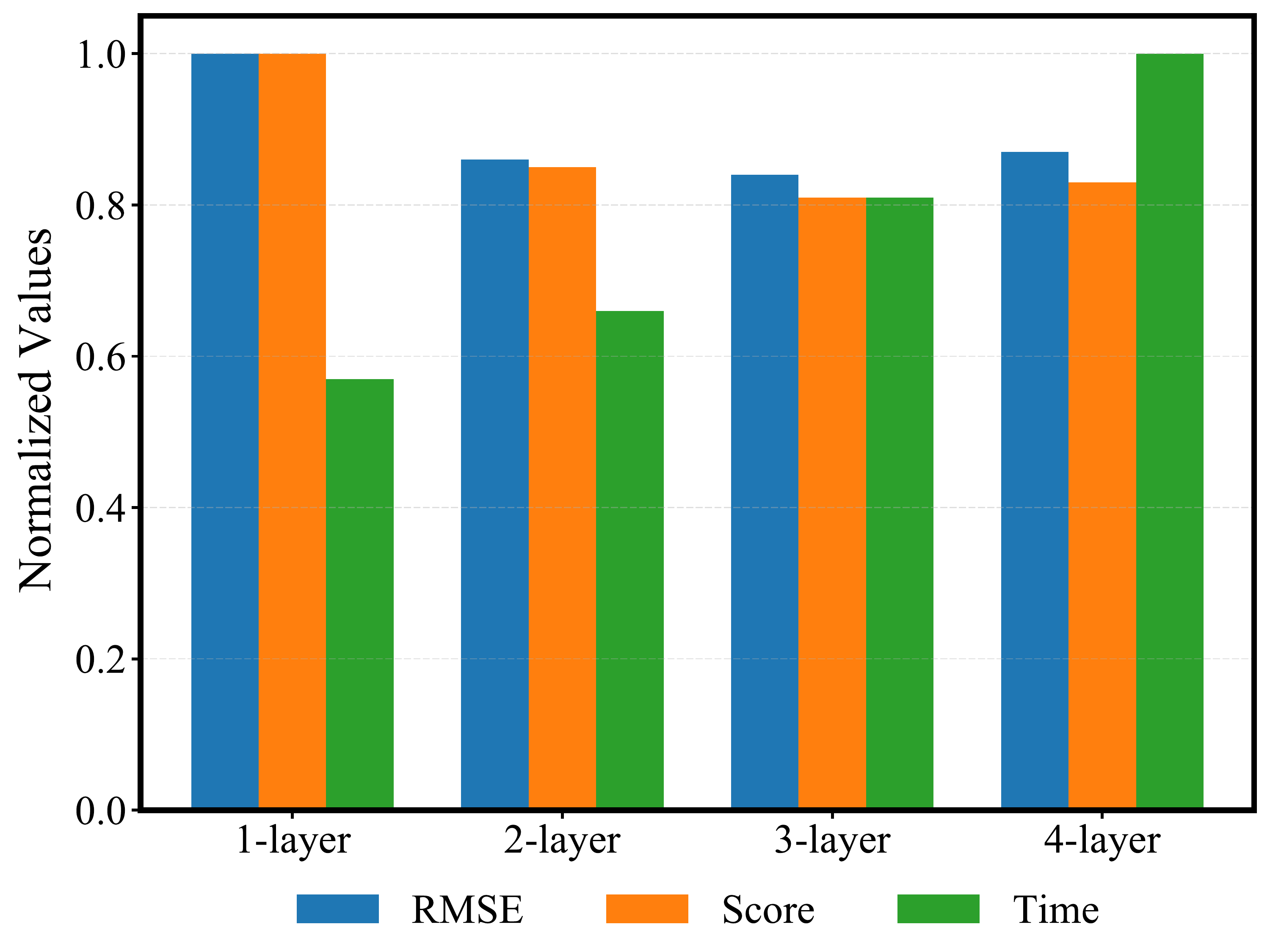}
	\caption{The number of convolutional network layers and performance metrics in terms of RMSE, Score and Time}
	\label{CNN_layer}
\end{figure}

For the 1D CNN, the depth of hidden convolutional network layers strongly affects feature extraction ability. The CNN unit with more stacked network layers usually can learn more abstract and discriminative representations but demands higher training time consumption. To study the size dependence of the TDDN prediction performance and time consumption, one to four network layers in the CNN unit is tested by fixing all other hyperparameters. The experimental results on the dataset FD001 are shown in Figure~\ref{CNN_layer}. The performance metrics, RMSE and Score, decrease with the number of the network layers. Clearly, deeper network layers can more effectively capture temporal features in the degradation development. In Fig.~\ref{CNN_layer}, three network layers already demonstrate the best performance while four network layers provide little extra performance gain.
Therefore, three network layers in the 1D CNN is the optimal choice with the trade-off between prediction accuracy and training time consumption.

\captionsetup[figure]{justification=raggedright}
\begin{figure}[!b]
	\centering
	\includegraphics[width=0.8\textwidth]{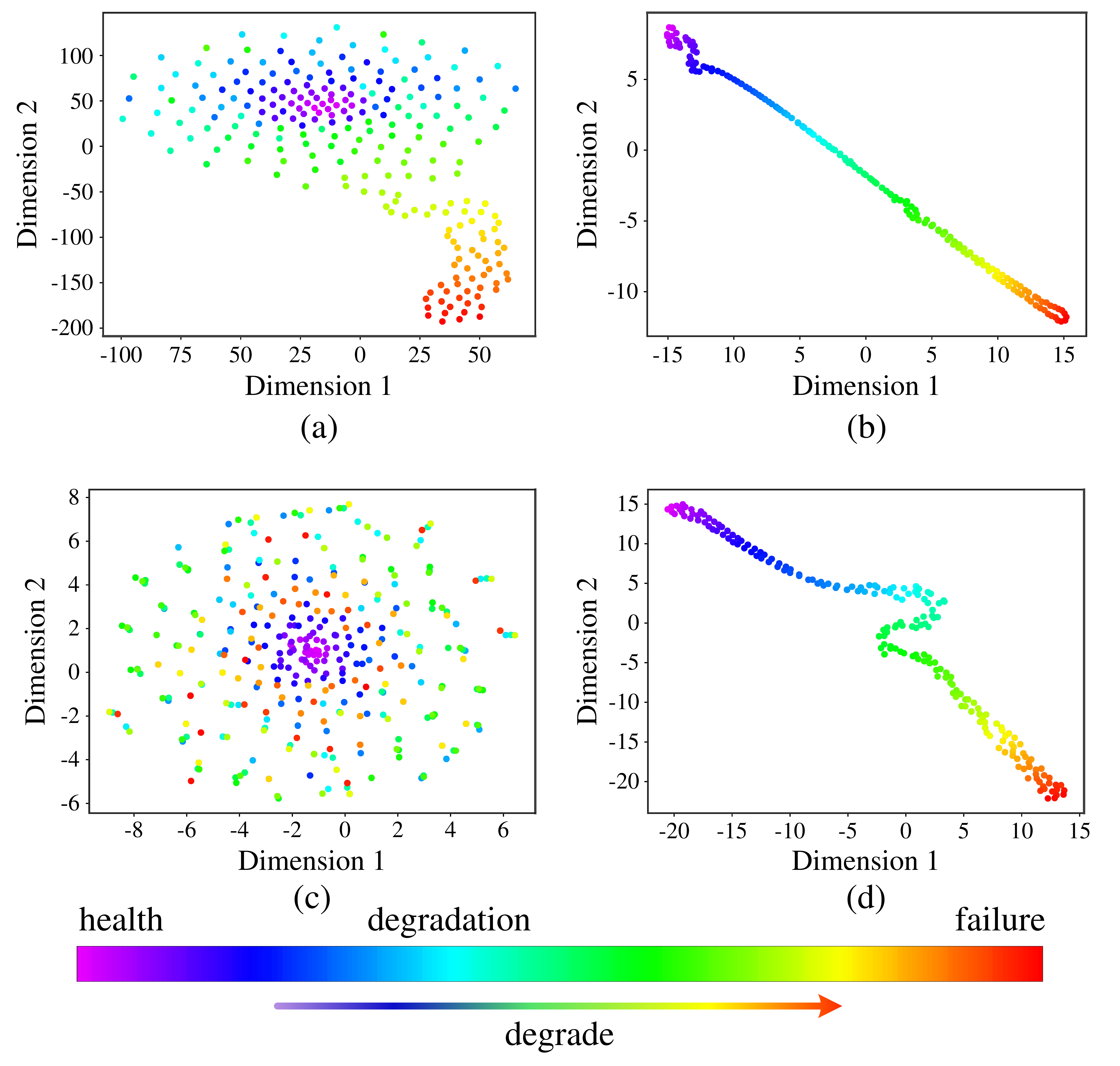}
	\caption{The t-SNE visualization of (a) raw sensor streaming data for engine 1 in the FD001 dataset, (b) temporal features for engine 1 in the FD001 dataset, (c) raw sensor streaming data for engine 1 in the FD004 dataset, (d)temporal features for the engine 1 in the FD004 dataset. 
	The dots of the t-SNE projection of the raw sensor streaming data in moving window $\mathbf{W}_j$ are visualized in (a) and (c). The dots of the t-SNE projection of the temporal features in moving window $\mathbf{W}_j$ are visualized in (b) and (d). The hotter the color is, the closer the engine is to the final failure.}
	\label{cluster}
\end{figure}
\subsection{Temporal Features and Degradation Development Trajectories}
\begin{figure}[!h]
	\centering
	\includegraphics[width=0.8\textwidth]{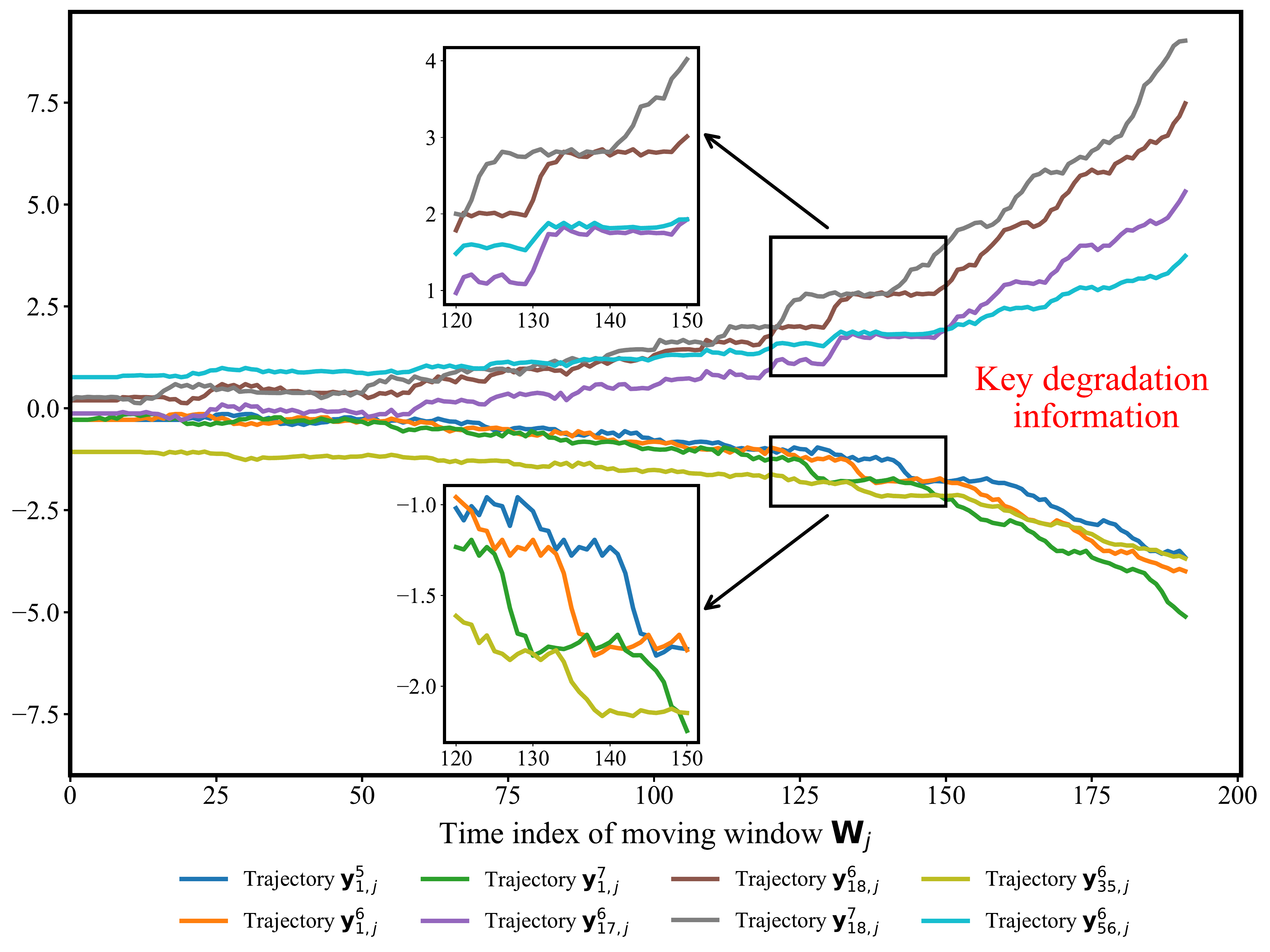}
	\caption{The trajectory of the $k$-th element of temporal feature $l$, $\mathbf{y}^k_{l,j}$ in the $j$-th moving window $\mathbf{W}_j$ for  engine 1 in the FD001 dataset.}
	\label{degradation_trajectory}
\end{figure}

To understand how the TDDN model works with its feature learning capability, t-stochastic neighbor embedding(t-SNE)\cite{van2008visualizing}, a nonlinear technique for unsupervised dimensionality reduction, is applied to visualize temporal features to gain more insights while preserving much of local structure in the high-dimensional space.
Fig.~\ref{cluster} illustrates the t-SNE projection of the raw sensor streaming data and temporal features extracted by the 1D CNN respectively at every time step $j$ in an engine degradation development.
Figs.~\ref{cluster} (a) and (c) shows the t-SNE dots of the raw sensor streaming data are highly mixed together and do not form any degradation trajectories because of high variance of the raw sensor streaming data as shown in Fig.~\ref{sensorFD001} and ~\ref{sensorFD004}.
In contrast, it is observed in Fig.~\ref{cluster} (b) and (d) that the t-SNE dots of the representative temporal features for moving window $\mathbf{W}_j$ form a clear degradation development trajectory, which is critical to improving the prediction performance of the TDDN model, particularly in the complex conditions of the FD004 dataset. The 1D CNN and optimal moving window can extract the degradation-related temporal features.
To further understand the degradation development trajectories in detail, the elements in eight temporal features extracted by the 1D CNN are plotted in Fig~\ref{degradation_trajectory}. Compared with the raw sensor streaming data in Fig.~\ref{sensorFD001}, the temporal features have clear trends and smaller fluctuation due to the 1D CNN filtering. The turning points in Fig.~\ref{degradation_trajectory} represent the essential degradation information in the middle degradation stage, which has a drastic change in a small time range. Therefore, the 1D CNN can learn degradation-related temporal features effectively and robustly.

\subsection{Attention Weights and Degradation Stages}
\begin{figure}[!t]
	\centering
	\subfigure[FD001]{
		\includegraphics[width=0.49\textwidth]{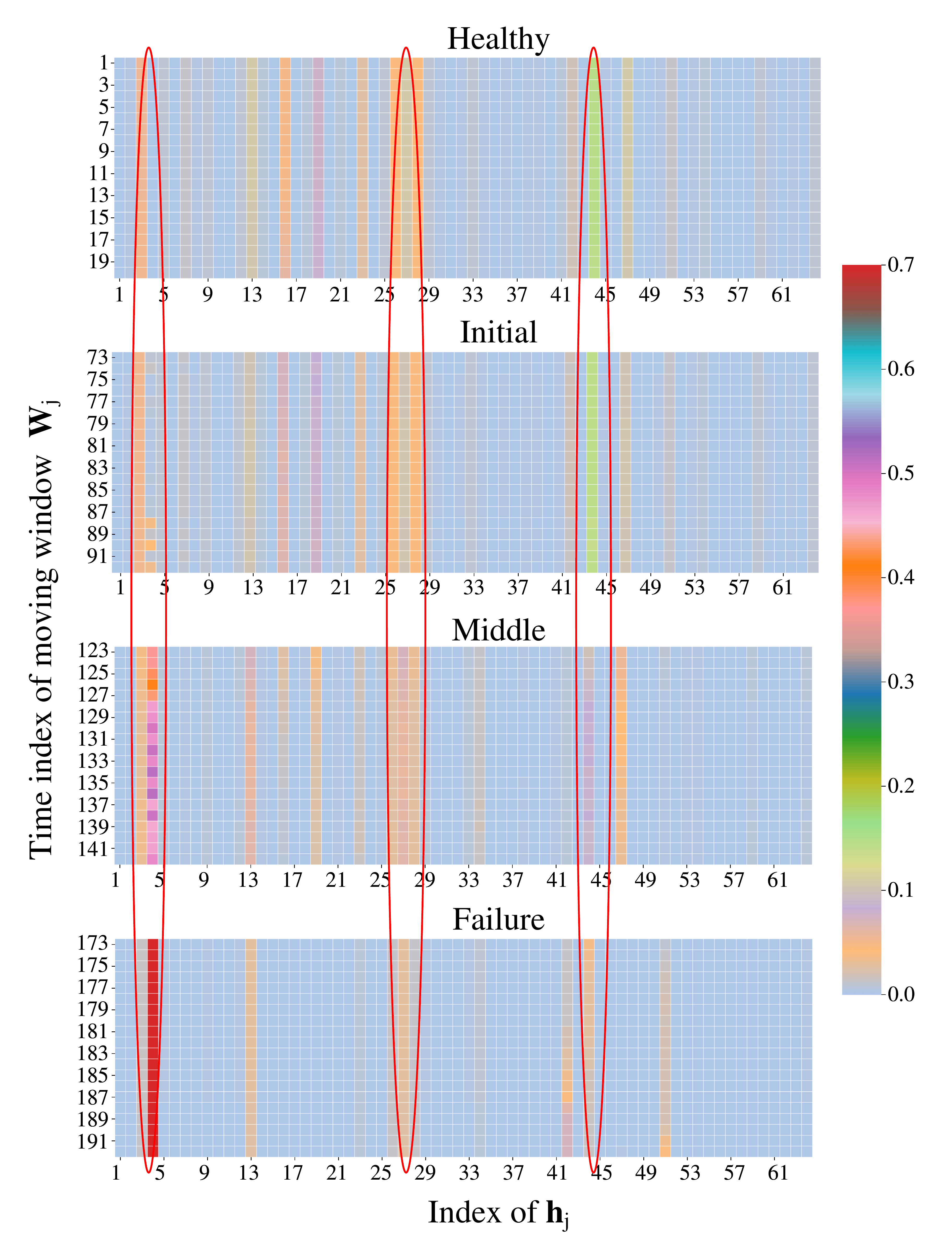}}
	\subfigure[FD004]{
		\includegraphics[width=0.49\textwidth]{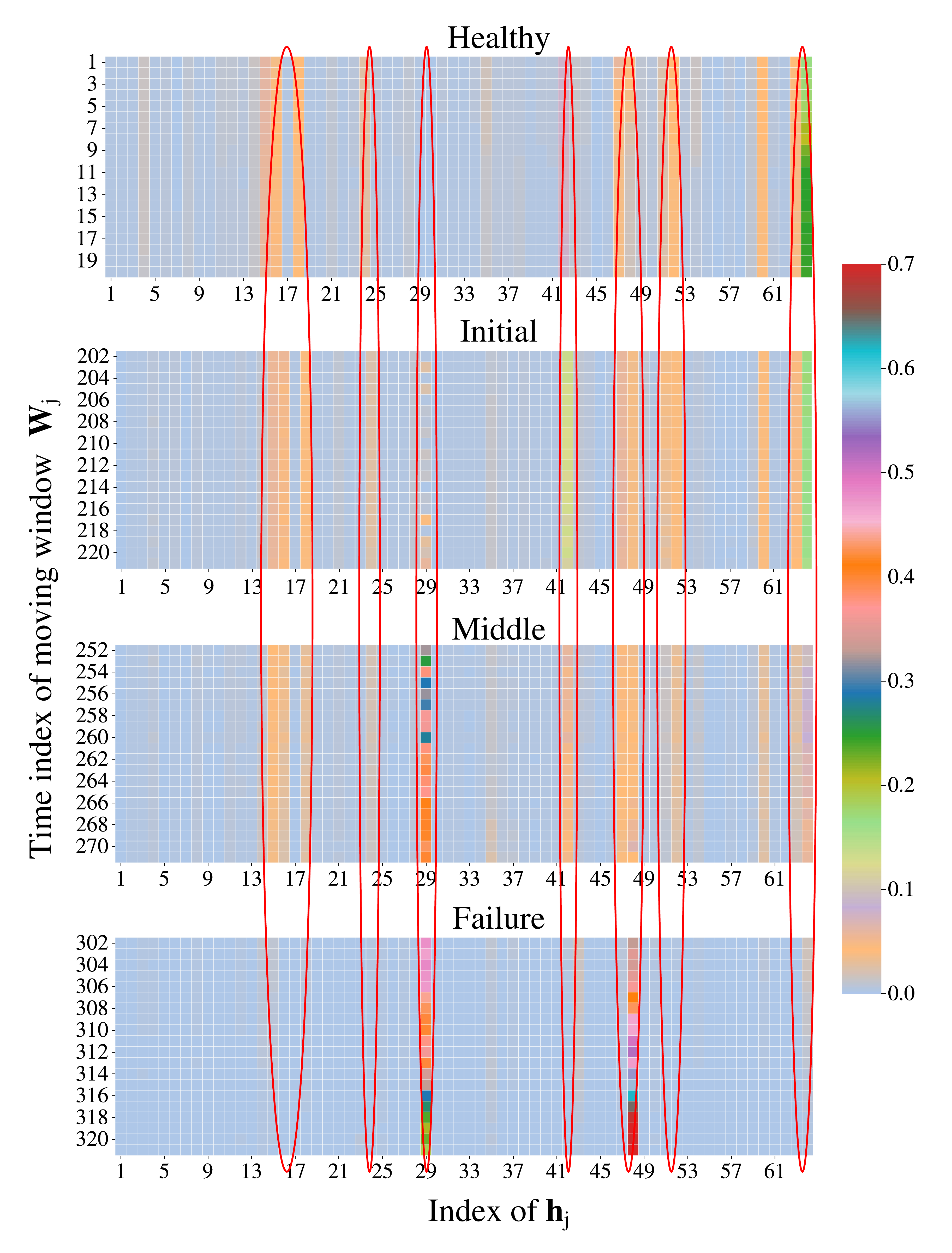}}
	\caption{The evolution of the attention weight $\lambda_{j}$ in moving window $\mathbf{W}_j$ for each abstract feature $\mathbf{h}_j$ for (a)  engine 1 in the FD001 dataset, (b)  engine 1 in the FD004 dataset. The twenty moving windows in four degradation stages are selected corresponding to the healthy stage, initial degradation stage, middle degradation stage, and failure stage. These stages are orderly shown in each picture from top to down.}
	\label{attention}
\end{figure}

\begin{figure}[!t]
	\centering
	\includegraphics[width=0.7\textwidth]{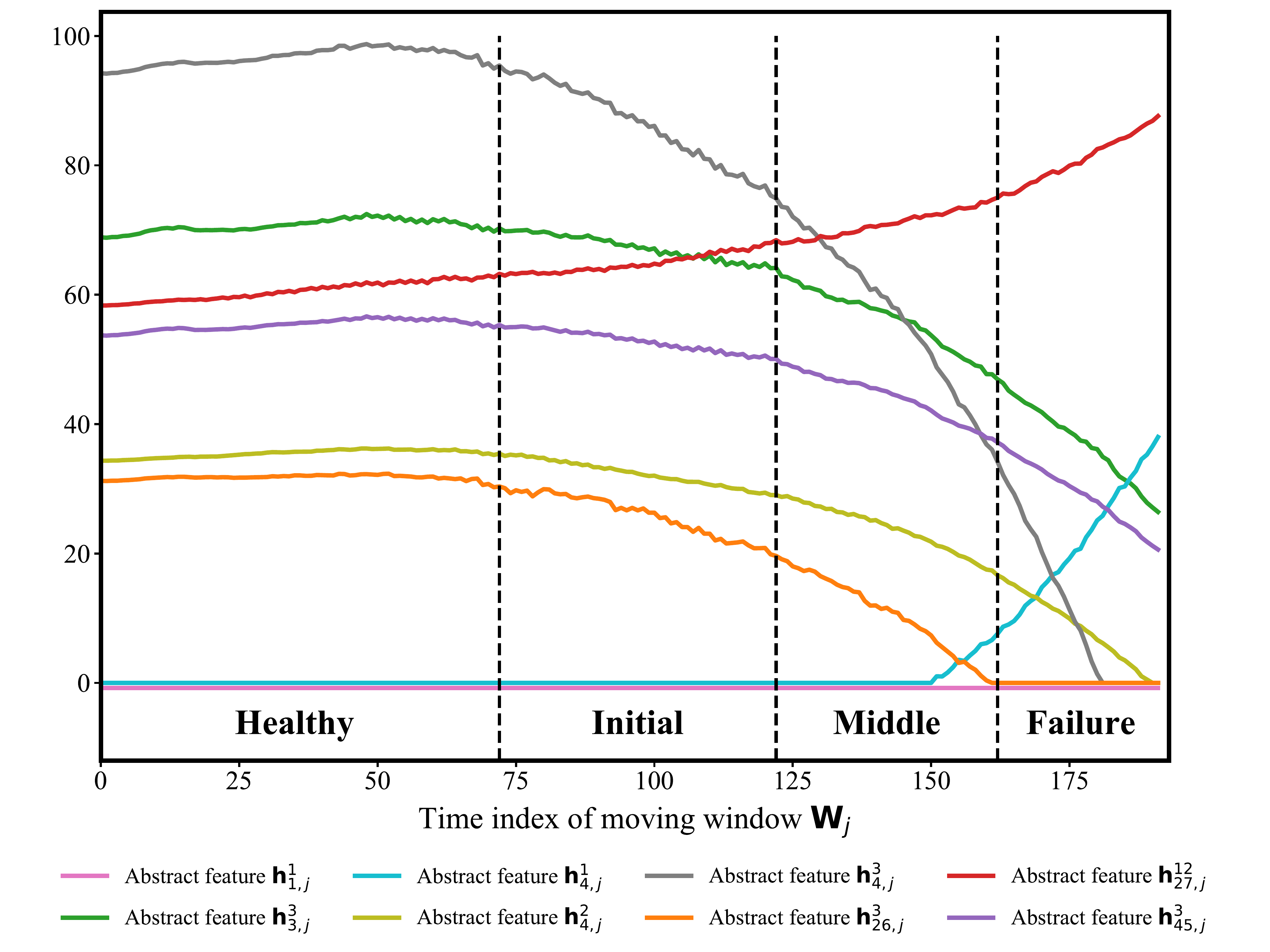}
	\caption{The trajectory of the $k$-th element of abstract feature $i$, $\mathbf{h}^{k}_{i,j}$ in moving window $\mathbf{W}_j$ for the engine 1 in the FD001 dataset. }
	\label{FD001_abstract_trajectory}
\end{figure}

\begin{figure}[!h]
	\centering
	\includegraphics[width=0.7\textwidth]{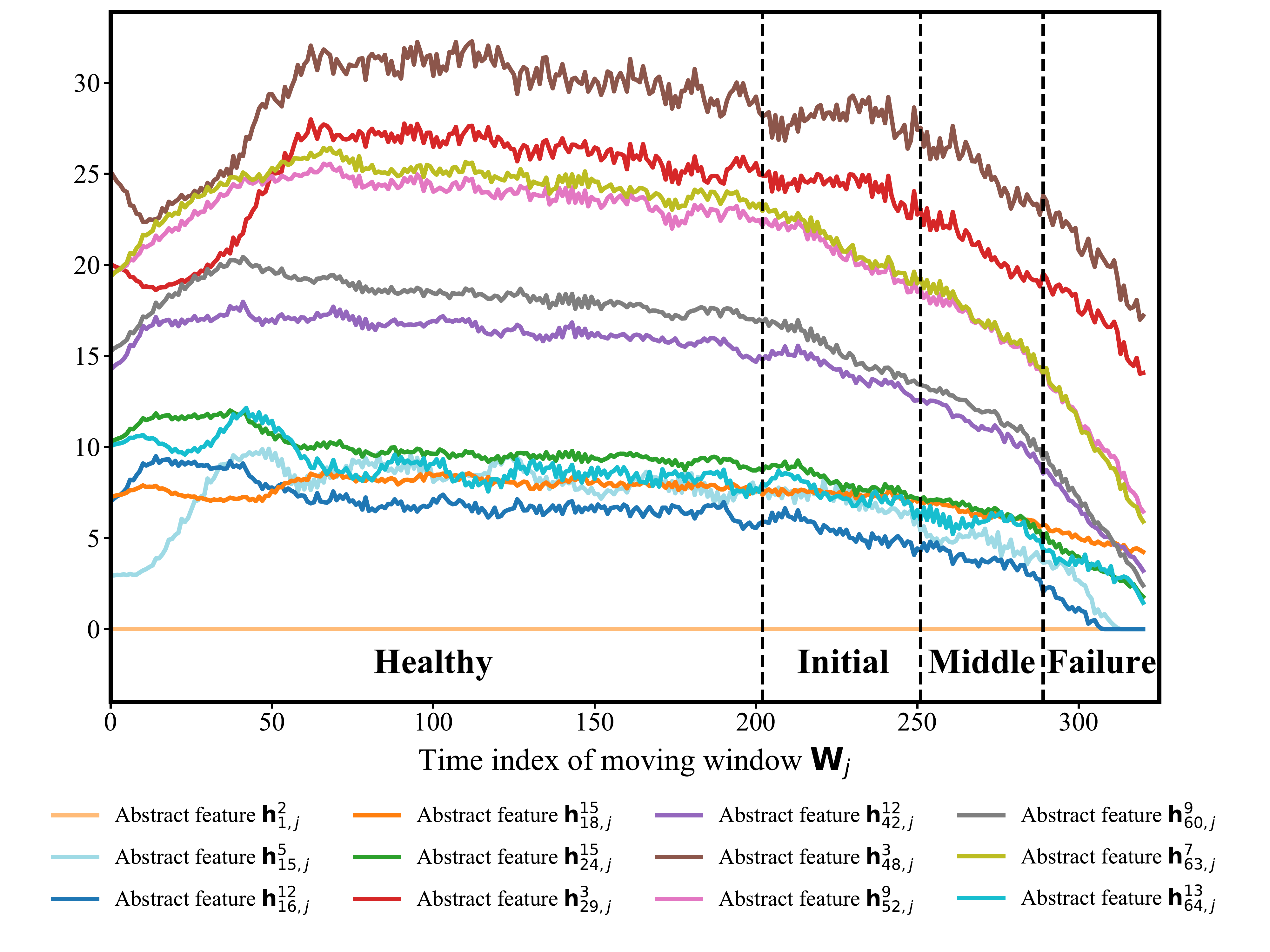}
	\caption{The trajectory of the $k$-th element of abstract feature $i$, $\mathbf{h}^{k}_{i,j}$ in moving window $\mathbf{W}_j$ for the engine  1 in the FD004 dataset}
	\label{FD004_abstract_trajectory}
\end{figure}

To better understand how the attention mechanism describes the degradation development in the TDDN model, the evolution of attention weight $\lambda_i$ for the abstract feature $\mathbf{h}_i$ in the trained TDDN model is visualized in Fig.~\ref{attention}. It is found that not all attention weights of the abstract features evolve with moving window $\mathbf{W}_j$. Some abstract features demonstrate no attention weight changing at all. 
The attention weights of the abstract features evolve with windows. Under different operating conditions, the attention weights for engine 1 in the FD001 dataset show the different evolution patterns from engine 1 in the FD004 dataset.
These abstract features with the attention weights that evolve with the degradation development are essentially key abstract features to characterize the degradation development and contain the important information for the RUL prediction.
The elements in the abstract features $\mathbf{h}_i$ in Fig.~\ref{FD001_abstract_trajectory} and Fig.~\ref{FD004_abstract_trajectory} have clear increasing or decreasing trend with moving windows. However, some elements in the abstract features do not change at all, and some only change at the failure stage. 
The underlying dynamics of the degradation will have dramatic changes.
The attention mechanism accurately captures the degradation patterns. The abstract features with higher attention weights in moving windows are crucial to the degradation development trajectories. Therefore, the attention mechanism can effectively focus on fault characteristics and improve the accuracy of RUL prediction.

\section{Conclusions}\label{discussion}
Combining the advantages of 1D CNN and the attention mechanism, the end-to-end TDDN deep learning framework is proposed to predict the turbofan engine RUL. Experiments are performed on a benchmark C-MAPSS dataset to evaluate the performance of the TDDN model. 
Compared with the performance metrics of existing deep learning methods on the C-MAPSS dataset, the TDDN model achieves the best RUL prediction.
Furthermore, since the 1D CNN and attention mechanism can learn degradation-related features, the TDDN model also has robust performance. Therefore, it is well suited for machinery prognostic problems in complex conditions.
Moreover, thanks to the learned monotonic-degradation temporal features and key fault characteristics capturing ability, the proposed method can also be used to find suitable maintenance decisions for industrial machinery. Furthermore, the effects of crucial hyperparameters are investigated. Moving window size is vital to catch the degradation development reflected in the fluctuating multivariate time series streaming data. 
1D CNN can automatically extract degradation-related temporal features from sensor streaming data and significantly enhance the feature-learning ability of the TDDN model. Meanwhile, the attention mechanism effectively identifies the underlying degradation development and captures key fault characteristics to improve prediction performance.

Despite promising prediction performance, the TDDN model accuracy significantly relies on a high-quality labeled dataset. Due to distributed sensors at different sites, the transmission of sensor streaming data to the server-side increases transmission costs and potential security issues in actual applications due to sensors distributed at different sites. Hence, the sensor data stakeholders are unwilling to share data for maintenance purposes, and thus sensor streaming data will be only available on local devices. The federated learning frameworks can help solve the issue. The federated learning aims to protect data privacy by enabling clients to collaboratively build machine learning models without sharing their data by training machine learning model locally and transmitting model parameters only. The machine learning model can be cooperatively learned without directly sharing sensitive streaming data. In the context of the Internet of Things (IoT), the TDDN model can be extended in the framework of federated learning to enhance data privacy and reduce data transmitting costs. The TDDN model can be effectively used in the distributed industrial machinery for the RUL prediction.

\section*{Acknowledgments}
Xin Chen acknowledges the funding support from the National Natural Science Foundation of China under grant No. 21773182 and the support of HPC Platform, Xi’an Jiaotong University.

\bibliography{cas-refs}
\end{document}